\documentclass{article}

\usepackage{microtype}
\usepackage{graphicx}
\usepackage{booktabs} 

\usepackage{hyperref}

\usepackage[accepted]{icml2023}

\usepackage{amsmath}
\usepackage{amssymb}
\usepackage{mathtools}
\usepackage{amsthm}

\usepackage{url}
\graphicspath{ {./images/} }
\usepackage{subcaption}
\usepackage{tikz}
\usepackage{float}
\usepackage[ruled, vlined, linesnumbered]{algorithm2e}
\usepackage{multirow,tabularx}
\usepackage{xspace}
\usepackage{alltt}
\usepackage{verbatim}
\usepackage{nicefrac}
\usepackage{enumitem}
\setlist[itemize]{
  noitemsep, topsep=0pt, label=$\blacktriangleright$, leftmargin=*,
}

\usepackage{amsmath,amsfonts,bm}

\newcommand{\newterm}[1]{{\bf #1}}

\def\figref#1{figure~\ref{#1}}
\def\Figref#1{Figure~\ref{#1}}

\def\eqref#1{equation~\ref{#1}}

\def\algref#1{algorithm~\ref{#1}}
\def\Algref#1{Algorithm~\ref{#1}}

\def\1{\bm{1}}
\newcommand{\train}{\mathcal{D}}

\def\eps{{\epsilon}}

\def\vm{{\bm{m}}}

\def\vv{{\bm{v}}}

\def\vx{{\bm{x}}}

\def\mR{{\bm{R}}}

\DeclareMathAlphabet{\mathsfit}{\encodingdefault}{\sfdefault}{m}{sl}
\SetMathAlphabet{\mathsfit}{bold}{\encodingdefault}{\sfdefault}{bx}{n}

\def\gM{{\mathcal{M}}}

\newcommand{\E}{\mathbb{E}}

\newcommand{\R}{\mathbb{R}}

\newcommand{\softmax}{\boldsymbol{\sigma}}

\DeclareMathOperator*{\argmin}{arg\,min}

\usepackage[
  disable,
  textsize=tiny,
  color=red!10
]{todonotes}

\usepackage{bold-extra}
\newcommand{\clam}{\texttt{\textbf{ClAM}}\xspace}

\def\brho{{\bm{\rho}}}
\def\bdelta{{\bm{\delta}}}

\newcommand{\up}{\textcolor{ForestGreen}{$\blacktriangle$}}
\newcommand{\down}{\textcolor{BrickRed}{$\blacktriangledown$}}

\newcommand{\txsps}[1]{${}^{\text{#1}}$}

\renewcommand{\paragraph}[1]{{\bf #1}}

\usepackage{multicol}
\usepackage{tcolorbox}
\theoremstyle{plain}
\newtheorem{theorem}{Theorem}[section]
\newtheorem{proposition}[theorem]{Proposition}
\newtheorem{lemma}[theorem]{Lemma}
\newtheorem{corollary}[theorem]{Corollary}
\theoremstyle{definition}

\theoremstyle{remark}

\icmltitlerunning{End-to-end Differentiable Clustering with Associative Memories}

\begin{document}

\twocolumn[
\icmltitle{End-to-end Differentiable Clustering with Associative Memories}



\icmlsetsymbol{equal}{*}
\begin{icmlauthorlist}
\icmlauthor{Bishwajit Saha}{yyy}
\icmlauthor{Dmitry Krotov}{MIT-IBM}
\icmlauthor{Mohammed J. Zaki}{yyy}
\icmlauthor{Parikshit Ram}{MIT-IBM,IBM}
\end{icmlauthorlist}
\icmlaffiliation{yyy}{CS Department, Rensselaer Polytechnic Institute, Troy, NY, USA}
\icmlaffiliation{MIT-IBM}{MIT-IBM Watson AI Lab, IBM Research, Cambridge, MA, USA}
\icmlaffiliation{IBM}{IBM Research, Yorktown Heights, NY, USA}
\icmlcorrespondingauthor{Bishwajit Saha}{sahab@rpi.edu}
%
\icmlkeywords{Machine Learning, ICML}
\vskip 0.3in
]
%
%
%
\printAffiliationsAndNotice{}  
%
\begin{abstract}
Clustering is a widely used unsupervised learning technique involving an intensive discrete optimization problem. Associative Memory models or AMs are differentiable neural networks defining a recursive dynamical system, which have been integrated with various deep learning architectures. We uncover a novel connection between the AM dynamics and the inherent discrete assignment necessary in clustering to propose {\em a novel unconstrained continuous relaxation of the discrete clustering problem}, enabling end-to-end differentiable {\bf cl}ustering with {\bf AM}, dubbed \clam. Leveraging the pattern completion ability of AMs, we further develop a novel self-supervised clustering loss. Our evaluations on varied datasets demonstrate that \clam benefits from the self-supervision, and significantly improves upon both the traditional Lloyd's $k$-means algorithm, and more recent continuous clustering relaxations (by upto 60\% in terms of the Silhouette Coefficient). 
\end{abstract}\vspace{-0.3in}
\section{Introduction}\label{sec:intro} 
Clustering is considered one of the most fundamental \newterm{unsupervised} techniques to identify the structure of large volumes of data. Based on the similarity characteristics, clustering performs a structured division of data into groups~\citep{xu2005survey}.
This is often considered the very first step in exploratory data analysis~\citep{saxena2017review, bijuraj2013clustering}. Various formulations and algorithms have been studied to identify an efficient clustering of the data. Among them $k$-means~\citep{macqueen1967classification}, Spectral Clustering~\citep{donath1973lower}, Hierarchical Clustering~\citep{johnson1967hierarchical}, Density-based Clustering~\citep{ester1996density}, and Expectation Maximization~\citep{dempster1977maximum} have been widely used. However, these formulations are computationally expensive involving an intensive combinatorial task: for example, exact $k$-means is NP-hard~\citep{dasgupta2008hardness}, though approximations can be efficient. Spectral, hierarchical and density-based clustering are computationally very expensive with a naive complexity between quadratic and cubic in the number of data points. One challenge of these formulations is the inability to leverage a differentiable algorithm -- they are inherently discrete.

There has been recent interest in clustering in an end-to-end differentiable manner. \newterm{Deep clustering} techniques combine representation learning and clustering and try to learn in a differentiable manner by leveraging some \newterm{continuous relaxation} of the discrete assignment necessary in clustering~\citep{ren2022deep, zhou2022comprehensive}.
This relaxation replaces the discrete assignment with {\em partial cluster assignments}, 
which violates a fundamental premise of clustering -- each point belongs to only one cluster. 
With this in mind, the ``sum-of-norms'' form of the $k$-means objective allows differentiable clustering, but requires quadratic time (in the number of points) to perform the final cluster assignments~\citep{panahi2017clustering}.
Non-negative matrix factorization with structured sparsity has also been considered but requires the repeated alternating solution of two large constrained least squares problems~\citep{kim2008sparse}.
To the best of our knowledge, {\em there is no differentiable clustering scheme that can seamlessly leverage the stochastic gradient descent or SGD~\citep{nemirovski2009robust} based optimization frameworks~\citep{duchi2011adaptive, kingma2014adam} for unconstrained optimization and yet maintain the inherent discrete nature of clustering}. There is also a problem with multiple local minima, which lead to suboptimal solutions with discrete algorithms. Beyond efficiency, SGD is known to be capable of escaping local minima and can lead to models with better generalization~\citep{hardt2016train}, and we believe that the problem of clustering can benefit from utilizing SGD based solutions. To that end, we look for ideas in a very distant field of associative memories.

Recently, traditional \newterm{associative memory} or AM  models~\citep{hopfield1982neural, hopfield1984neurons} have been reformulated to significantly increase their memory storage capacity and integrated with modern deep learning techniques~\citep{krotov2016dense, ramsauer2020hopfield, krotov2021large, krotov2021hierarchical}. These novel models, called Dense Associative Memories, are fully differentiable systems capable of storing a large number of multi-dimensional vectors, called patterns or ``memories'', in their synaptic weights. They can be trained in an end-to-end fully differentiable setting using the backpropagation algorithm, typically with a self-supervised loss.

\begin{figure}[t]
\centering
\begin{subfigure}{0.225\columnwidth}
\centering
\centerline{\includegraphics[width=\textwidth]{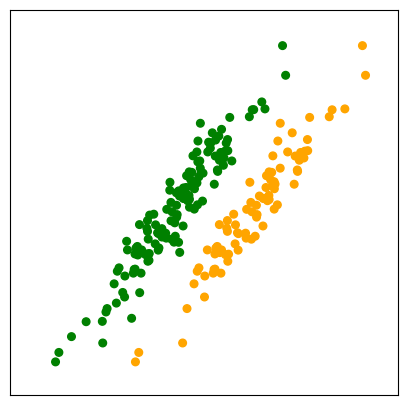}}
\caption{True}
\label{fig:original}
\end{subfigure}
~
\begin{subfigure}{0.225\columnwidth}
\centering
\centerline{\includegraphics[width=\textwidth]{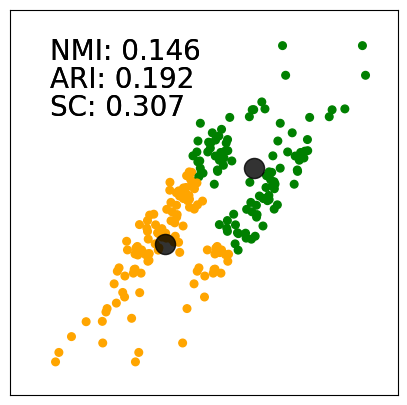}}
\caption{Lloyd's}
\label{fig:kmeans}
\end{subfigure}
~
\begin{subfigure}{0.225\columnwidth}
\centering
\centerline{\includegraphics[width=\textwidth]{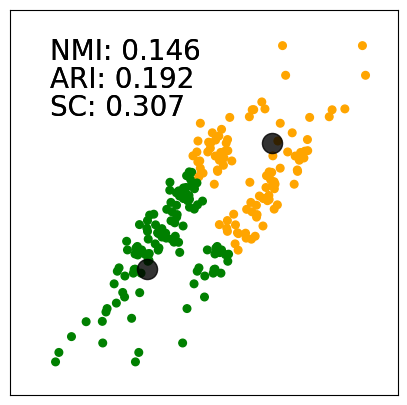}}
\caption{DEC}
\label{fig:skmeans}
\end{subfigure}
~
\begin{subfigure}{0.225\columnwidth}
\centering
\centerline{\includegraphics[width=\textwidth]{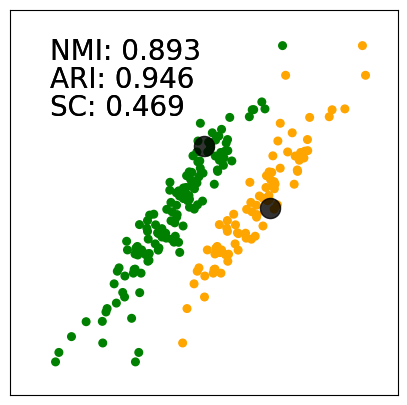}}
\caption{\clam}
\label{fig:clam}
\end{subfigure}
\vspace{-0.1in}
\caption{{\bf Clustering with \clam.} Two clusters (\figref{fig:original}), and solutions found by $k$-means~\citep{lloyd1982least} (\figref{fig:kmeans}), DEC relaxation~\citep{xie2016unsupervised} of $k$-means (\figref{fig:skmeans}), and our proposed end-to-end differentiable SGD-based \clam (\figref{fig:clam}).
The black dots indicate the learned prototypes. {\em \clam discovers the ground-truth clusters while the baselines cannot}. See experimental details in \S \ref{sec:eval}.%
\vspace{-0.2in}}
\label{fig:teaser:toy-result}
\end{figure}

We believe that this ability to learn synaptic weights of the AM in an end-to-end differentiable manner together with the discrete assignment (association) of each data point to exactly one memory makes AM uniquely suited for the task of differentiable clustering. We present a simple result in \figref{fig:teaser:toy-result}, where standard prototype-based clustering schemes (discrete and differentiable) are unable to find the right clusters, but our proposed AM based scheme is successful. Specifically, we make the following contributions:
\begin{itemize}
\item We develop a flexible mathematical framework for {\bf cl}ustering with {\bf AM} or \clam, which is a {\em novel continuous unconstrained relaxation} of the discrete optimization problem of clustering that allows for clustering in an {\bf end-to-end differentiable manner} while maintaining the discrete cluster assignment throughout the training, with linear time cluster assignment.
\item We leverage the pattern completion capabilities of AMs to develop a {\bf differentiable self-supervised loss} that improves the clustering quality.
\item We empirically demonstrate that \clam is able to consistently improve upon $k$-means by upto 60\%, while being competitive to spectral and agglomerative clustering, and producing insightful interpretations.
\end{itemize}

\section{Related work} \label{sec:related_work}
\paragraph{Discrete clustering algorithms.}
Clustering is an inherently hard combinatorial optimization problem. Prototype based clustering formulations such as the $k$-means clustering is NP-hard, even for $k=2$~\citep{dasgupta2008hardness}. However, the widely used Lloyd's algorithm or Voronoi iteration~\citep{lloyd1982least} can quite efficiently find locally optimal solutions, which can be improved upon via multiple restarts and careful seeding~\citep{vassilvitskii2006k}. It is an intuitive discrete algorithm, alternating between (i)~assigning points to clusters based on the Voronoi partition~\citep{aurenhammer1991voronoi} induced by the current prototypes, and (ii)~updating prototypes based on the current cluster assignments. However, the discrete nature of this scheme makes it hard to escape any local minimum.
{\em We believe that SGD-based prototype clustering will improve the quality of the clusters by being able to escape local minima, while also inheriting the computational efficiency of SGD}.
The spectral clustering formulation~\citep{donath1973lower, von2007tutorial} utilizes the eigen-spectrum of the pairwise similarity or affinity of the points (represented as a graph Laplacian) to partition the data into ``connected components'' but does not explicitly extract prototypes for each cluster. This allows spectral clustering to partition the data in ways disallowed by Voronoi partitions. However, it naively requires quadratic time (in the number of points) to generate the Laplacian, and cubic time for the spectral decomposition.
Hierarchical clustering~\citep{johnson1967hierarchical} finds successive clusters based on previously established clusters, dividing  them in a top-down fashion or combining in a bottom-up manner. These discrete hierarchical algorithms naively scale cubically in the number of points, although some forms of bottom-up agglomerative schemes can be shown to scale in quadratic~\citep{sibson1973slink} or even sub-quadratic time~\citep{march2010fast}  leveraging dual-tree algorithms~\citep{ram2009linear, curtin2013tree, curtin2015plug}.

\paragraph{Density-based clustering.}
Schemes, such as the popular DBSCAN~\citep{ester1996density} and SNN~\citep{aguilar2001snn}, 
do not view clustering as a discrete optimization problem but as the problem of finding the \newterm{modes} in the data distribution, allowing for arbitrary shaped clusters with robustness to noise and outliers~\citep{sander1998density}. However, multi-dimensional density estimation is a challenging task, especially with nonparametric methods~\citep{silverman1986density, ram2011density}, leading to the use of parametric models such as Gaussian Mixture Models~\citep{reynolds2009gaussian} (which can be estimated from the data with Expectation-Maximization~\citep{dempster1977maximum}).

\paragraph{Differentiable \& deep clustering.}
Differentiability is of critical interest in deep clustering where we wish  to simultaneously learn a latent representation of the points and perform clustering in that latent space~\citep{ren2022deep, zhou2022comprehensive}. Existing differentiable clustering formulations such as the sum-of-norms~\citep{panahi2017clustering} or the non-negative matrix factorization~\citep{kim2008sparse} based ones are essentially solving constrained optimization and do not directly fit into an end-to-end differentiable deep learning pipeline. Hence, various schemes utilize some form of soft clustering formulation such as fuzzy $k$-means~\citep{bezdek1984fcm}.
\citet{xie2016unsupervised} proposed a novel probabilistic $k$-means formulation inspired by t-SNE~\citep{van2008visualizing} to perform differentiable clustering (DEC) on representations from a pretrained autoencoder (AE), while \citet{guo2017improved} (IDEC) showed gains from jointly learning representations and clusters.
For representation learning in deep clustering of images, \citet{song2013auto}, \citet{xie2016unsupervised}, and \citet{yang2017towards} used stacked autoencoders, and \citet{guo2017deep} (DCEC) showed further improvements with convolutional autoencoders (CAE).
\citet{chazan2019deep} utilized multiple AEs, one per cluster, but the (soft) cluster assignment was performed with fuzzy $k$-means. \citet{cai2022unsupervised} added a ``focal loss'' to the DEC relaxation to penalize non-discrete assignments, while also enhancing the representation learning. Deep subspace clustering~\citep{peng2016deep, ji2017deep, zhang2019self} relaxes the combinatorial spectral clustering problem (again with partial cluster assignments), and utilizes ``self-expressive'' layers to simultaneously learn the necessary Laplacian (affinity matrix) and the clusters in a differentiable manner. 
Thus, deep clustering generally uses some probabilistic partial assignment of points to multiple clusters, deviating from the discrete nature of original $k$-means.
Furthermore, these schemes often leverage pretrained AEs and/or $k$-means via \citet{lloyd1982least} to initialize the network parameters. {\em Our proposed \clam maintains the discrete assignment nature of clustering, and works well without any special seeding}. 

\paragraph{Associative Memory (AM).}
This is a neural network that can store a set of multidimensional vectors -- {\bf memories} -- as fixed point {\bf attractor states} of a recurrent dynamical system. It is designed to {\bf associate} the initial state (presented to the system) to the final state at a fixed point (a memory), thereby defining disjoint \newterm{basins of attractions} or partitions of the data domain, and thus, can represent a clustering.
This network was mathematically formalized as the classical Hopfield Network~\citep{hopfield1982neural}, but is known to  have limited memory capacity, being able to only store $\approx 0.14d$ random memories in a $d$ dimensional data domain~\citep{amit1985storing, mceliece1987capacity}. For correlated data,
the capacity is even smaller, and often results in one fixed point attracting the entire dataset into a single basin.
This behavior is problematic for clustering, since the number of clusters --  the number of stable fixed points -- should be decoupled from the data dimensionality.
\citet{krotov2016dense} proposed Modern Hopfield Network or Dense AM by introducing rapidly growing non-linearities -- activation functions -- into the dynamical system, allowing for a denser arrangement of memories, and super-linear (in $d$) memory capacity. For certain activation functions, Dense AMs have power law or even exponential capacity~\citep{krotov2016dense, demircigil2017model,ramsauer2020hopfield,lucibello2023exponential}. 
\citet{ramsauer2020hopfield} demonstrated that the attention mechanism in transformers~\citep{vaswani2017attention} is a special limiting case of Dense AMs with the softmax activation. Dense AMs have also been used to describe the entire transformer block \citep{hoover2023energy}, as well as integrated in sophisticated energy-based neural architectures \citep{hooveruniversal}. See \citet{krotov2023new} for a review of these results. Dense AMs have also been recently studied with setwise connections \citep{burns2023simplicial}, as well as applied to hetero-associative settings \cite{liang2022modern}. 
In our work, we will show that the recurrent dynamics and the large memory capacity of Dense AMs make them uniquely suitable for clustering.
\section{Associative Memories for Clustering} \label{sec:clam}
In this section, we present (i)~the mathematical framework for \clam based on AMs, (ii)~motivate its suitability for clustering, and (iii)~present a way of learning the memories for good clustering. We put all this together in our novel prototype-based clustering algorithm, \clam.

\subsection{Associative memories: Mathematical Framework}\label{sec:clam:am}

In a $d$-dimensional Euclidean space, consider $M$ memories $\brho_\mu \in \R^d, \mu \in [M] \triangleq \{ 1, \ldots, M \}$ (we will discuss later how the memories are learned). The critical aspects of this mathematical framework are the \newterm{energy function} and the \newterm{attractor dynamics}~\citep{krotov2021large, millidge2022universal}.
A suitable energy for clustering should be a continuous function of a point (or a particle) $\vv \in \R^d$.
Additionally, the energy should have $M$ local minima, corresponding to each memory. Finally, as a particle progressively approaches a memory, its energy should be primarily determined by that single memory, while the contribution of the remaining $M-1$ memories should be small.
An energy function satisfying these requirements is given by
\begin{equation}\label{eq:particle-energy}
  E(\vv) = -\frac{1}{\beta}\log
  \left(\sum\nolimits_{\mu\in[M]} \exp ( -\beta \| \brho_\mu - \vv \|^2 ) \right)
\end{equation}
with a scalar $\beta > 0$ interpreted as an inverse ``temperature''.
As $\beta$ grows, the $\exp(\cdot)$ in \eqref{eq:particle-energy} ensures that only the leading term remains significant, while the remaining $M-1$ terms are suppressed.
Thus, the entire energy function will be described by a parabola around the closest memory. The space will be partitioned into $M$ basins of attraction around each memory, and the shape of the energy function around each memory will be primarily defined by the closest memory, with small corrections from other memories.  

The attractor dynamics control how $\vv$ moves in the space over time, via $d\vv/dt$,
while ensuring that energy decreases.
That is $dE(\vv)/dt < 0$,
which ensures that a particle convergences to a local minimum -- a fixed point of the dynamics -- given the lower bounded energy.
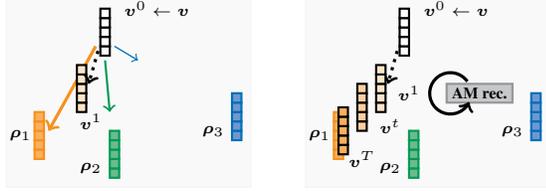
\begin{figure}[t]
\centering
\begin{subfigure}{0.45\columnwidth}
\centering
{\tiny
\begin{tikzpicture}[scale=0.125]
  \filldraw[color=Gray!10, fill=Gray!5, very thin] (0, 0) rectangle (26, 20);
  \foreach \i in {1,...,5}
  {
    \filldraw[color=BurntOrange, fill=BurntOrange!70, thick] (3, 2+\i) rectangle (4, 3+\i);
  }
  \node (r1) at (1.5, 5.5) {$\brho_1$};
  \foreach \i in {1,...,5}
  {
    \filldraw[color=ForestGreen, fill=ForestGreen!60, thick] (11, 0+\i) rectangle (12, 1+\i);
  }
  \node (r2) at (9, 2) {$\brho_2$};
  \foreach \i in {1,...,5}
  {
    \filldraw[color=NavyBlue, fill=NavyBlue!60, thick] (24, 4+\i) rectangle (25, 5+\i);
  }
  \node (r3) at (22, 6) {$\brho_3$};
  \foreach \i in {1,...,5}
  {
    \filldraw[color=black, fill=white, thick] (10, 13+\i) rectangle (11, 14+\i);
  }
  \node (vv) at (16, 19) {$\vv^{0} \gets \vv$};
  \draw[very thick, ->, BurntOrange] (9.5, 15) -- (4.5, 6);
  \draw[thick, ->, ForestGreen] (10.5, 13.5) -- (11, 8);
  \draw[thin , ->, NavyBlue, line width=0.5pt] (11.5, 15) -- (14, 13.5);
  \foreach \i in {1,...,5}
  {
    \filldraw[color=black, fill=BurntOrange!20, thick] (7.5, 7+\i) rectangle (8.5, 8+\i);
  }
  \node (vv1) at (9, 7) {$\vv^{1}$};
  \draw[very thick, dotted, ->, black] (9.75, 14.5) -- (8.75, 11);
\end{tikzpicture}
}
\caption{Attraction and update.}
\label{fig:am-dym:one-step}
\end{subfigure}
~
\begin{subfigure}{0.45\columnwidth}
\centering
{\tiny
\begin{tikzpicture}[scale=0.125]
  \filldraw[color=Gray!10, fill=Gray!5, very thin] (0, 0) rectangle (26, 20);
  \foreach \i in {1,...,5}
  {
    \filldraw[color=BurntOrange, fill=BurntOrange!70, thick] (3, 2+\i) rectangle (4, 3+\i);
  }
  \node (r1) at (1.5, 5.5) {$\brho_1$};
  \foreach \i in {1,...,5}
  {
    \filldraw[color=ForestGreen, fill=ForestGreen!60, thick] (11, 0+\i) rectangle (12, 1+\i);
  }
  \node (r2) at (9, 2) {$\brho_2$};
  \foreach \i in {1,...,5}
  {
    \filldraw[color=NavyBlue, fill=NavyBlue!60, thick] (24, 4+\i) rectangle (25, 5+\i);
  }
  \node (r3) at (22, 6) {$\brho_3$};
  \foreach \i in {1,...,5}
  {
    \filldraw[color=black, fill=white, thick] (10, 13+\i) rectangle (11, 14+\i);
  }
  \node (vv) at (16, 19) {$\vv^{0} \gets \vv$};
  \node (vv1) at (11, 10) {$\vv^{1}$};
  \draw[very thick, dotted, ->, black] (9.75, 14.5) -- (8.75, 11);
  \foreach \i in {1,...,5}
  {
    \filldraw[color=black, fill=BurntOrange!70, thick] (3.5, 2.5+\i) rectangle (4.5, 3.5+\i);
  }
  \node (vvT) at (6, 3) {$\vv^{T}$};
  \node (vvt) at (9, 6.5) {$\vv^{t}$};
  \filldraw[color=Gray, fill=Gray!50, thick] (15, 9) rectangle (22, 11);
  \node (amr) at (18.5, 10) [align=left] {\bf AM rec.};
  \draw [->, very thick] (17.2, 11.25) arc [start angle=35, end angle=335, x radius=2.2, y radius=2.2];
  \foreach \i in {1,...,5}
  {
    \filldraw[color=black, fill=BurntOrange!20, thick] (7.5, 7+\i) rectangle (8.5, 8+\i);
  }
  \foreach \i in {1,...,5}
  {
    \filldraw[color=black, fill=BurntOrange!40, thick] (5.5, 5+\i) rectangle (6.5, 6+\i);
  }  
\end{tikzpicture}
}
\caption{Recursion to a fixed point.}
\label{fig:am-dym:cvg}
\end{subfigure}
\vspace{-0.1in}
\caption{{\bf Attractor dynamics.} \Figref{fig:am-dym:one-step} visualizes the attraction (direction \& magnitude with solid colored arrows) of a particle $\vv^{0}$ toward each of the memories $\brho_1, \brho_2, \brho_3$, and the resulting update $\bdelta^{1}$ (black dotted arrow) from 
$\vv^{0} \to \vv^{1}$
(\eqref{eq:att-dyn-finite}). \Figref{fig:am-dym:cvg} visualizes the recursive application (AM recursion) of \eqref{eq:att-dyn-finite} to the particle 
$\vv^{0} \to \vv^{1} \to \cdots \to \vv^{t} \to \cdots$
till it converges to a memory $\vv^{t=T} \approx \brho_1$.%
\vspace{-0.2in}}
\label{fig:am-dym}
\end{figure}
The particle dynamics is 
described by gradient descent on the energy landscape:
\begin{equation}\label{eq:att-dyn}
\tau \frac{d\vv}{dt} = -\frac{1}{2} \nabla_\vv E =\sum_{\mu \in [M]} (\brho_\mu - \vv) \; \softmax(- \beta \| \brho_\mu - \vv \|^2 )
\end{equation}
where $\tau > 0$ is a characteristic \newterm{time constant}, describing how quickly the particle moves on the energy landscape, and $\softmax(\cdot)$ is the softmax function over the scaled distances to the memories, defined as $\softmax(z_\mu) = \nicefrac{\exp(z_\mu)}{\sum_{m=1}^M \exp(z_m)}$ for any $\mu\in [M]$.
This is visualized in \figref{fig:am-dym:one-step}.
This is guaranteed to reduce the energy since
%
\begin{equation}\label{eq:energy-decrease}
    \frac{dE(\vv)}{dt} \stackrel{\text{(a)}}{=} \nabla_\vv E(\vv) \cdot \frac{d\vv}{dt} \stackrel{\text{(b)}}{=} -2\tau \Big\|\frac{d\vv}{dt}\Big\|^2
    \stackrel{\text{(c)}}{\leq} 0
\end{equation}
where (a) is the chain rule, (b) follows from \eqref{eq:att-dyn}, and the equality in (c) implying local stationarity.
A valid update $\bdelta^{t+1}$ for the point $\vv$ from state $\vv^{t}$ to $\vv^{t+1} = \vv^{t} + \bdelta^{t+1}$ at a discrete time-step $t+1$ is via finite differences:%
\begin{align}\label{eq:att-dyn-finite}
\bdelta^{t+1}
=
\frac{dt}{\tau} \sum\nolimits_{\mu\in [M]} (\brho_\mu - \vv^{t} ) \; \softmax (-\beta \| \brho_\mu - \vv^{t} \|^2 )
\end{align}
Given a dataset of points $S$, for each point $\vv \in S$ one can corrupt it with some noise to produce a distorted point $\tilde\vv$. This serves as an initial state of the AM network $\vv^{0} \gets \tilde\vv$. The AM dynamics is defined by the learnable weights $\brho_\mu, \mu = 1, \ldots, M$, which also correspond to the fixed points of the dynamics (memories). The network evolves in time for $T$ recursions according to \eqref{eq:att-dyn-finite},
where $T$ is chosen to ensure sufficient convergence to a fixed point (see example in \figref{fig:am-dym:cvg}).
The final state $\vv^{T}$
is compared with the uncorrupted $\vv$ to define the loss function
\begin{equation}\label{eq:am-loss}
\mathcal{L} =  \sum\nolimits_{\vv \in S} \| \vv - \vv^{T} \|^2, \text{ where } \vv^{0} \gets \tilde\vv
\end{equation}
%
which is minimized with backpropagation through time with respect to the AM parameters $\brho_\mu$.
In the context of clustering, the AM model naturally induces a partition of the data space into non-overlapping basins of attraction, implicitly defines a hard cluster assignment as the fixed point in each basin, and is achieved through the fully continuous and differentiable dynamics of AM, allowing learning with standard deep learning frameworks, as we discuss next.
%
\subsection{AM as a differentiable discrete $\argmin$ solver} \label{sec:clam:mot}
Consider the original $k$-means objective with a dataset $S \subset \R^d$, where we learn $k$ prototypes $\mR \triangleq \{\brho_\mu, \mu \in [k]\}$ with $[k] \triangleq \{1, \ldots, k\}$ by solving the following problem:
\begin{equation}\label{eq:kmeans-obj}
\min_\mR
\sum_{\vx \in S} \| \vx - \brho_{\mu^\star_\vx} \|^2,
\text{\sf s.t. }
\mu^\star_\vx = \argmin_{\mu \in [k]} \| \vx - \brho_\mu \|^2
\end{equation}
%
The discrete selection of $\mu^\star_\vx$ (for each $\vx$) makes \eqref{eq:kmeans-obj} a combinatorial optimization problem that cannot directly be solved via (stochastic) gradient descent. A common continuous relaxation of this problem is as follows:
\begin{equation} \label{eq:soft-kmeans-obj}
\min_\mR
\sum\nolimits_{\vx \in S}
\sum\nolimits_{\mu \in [k]}
w_\mu(\vx)\;
\| \vx - \brho_\mu \|^2
\end{equation}
where (usually) $w_\mu(\vx) \in [0,1]$ 
and $\sum_{\mu \in [k]} w_\mu(\vx) = 1\, \forall \vx \in S$. Hence, these weights $\{ w_\mu(\vx), \mu \in [k]\}$ define a probability over the $k$ prototypes, and are designed to put the most weight on the closest prototype $\brho_{\mu^\star_\vx}$, and as small a weight as possible on the remaining prototypes.
For example, the softmax function $\softmax(\cdot)$ with distances to the prototypes has been used as
%
$w_\mu(\vx) = \softmax(-\gamma \|\vx - \brho_\mu\|^2)$,
%
where $\gamma > 0$ is a hyperparameter, and as $\gamma \to \infty$, $\sum_{\mu \in [k]} w_\mu(\vx)\| \; \vx - \brho_\mu\|^2 \to \| \vx - \brho_{\mu^\star_\vx} \|^2$. Other  weighting functions have been developed with similar properties~\citep{xie2016unsupervised},
and essentially utilize a weighted sum of distances to the learned prototypes,
resulting in something different in essence to the discrete assignment $\mu^\star_\vx$ in the original problem (\eqref{eq:kmeans-obj}). Another subtle point is that this {\bf soft weighted assignment} of any $\vx \in S$ across all prototypes {\bf at training time} does not match the {\bf hard cluster assignment} to the nearest prototype {\bf at inference}, introducing an incongruity between training and inference.

We propose a novel alternative continuous relaxation to the discrete $k$-means problem leveraging the AM dynamics (\S \ref{sec:clam:am}) that preserves the discrete assignment in the $k$-means objective. Given the prototypes (memories) $\mR = \{ \brho_\mu, \mu \in [k] \}$, the dynamics (\eqref{eq:att-dyn}) and the updates (\eqref{eq:att-dyn-finite}) ensure that any example (particle) $\vx \in \R^d$ will converge {\em to exactly one of the prototypes} $\brho_{\widehat\mu_\vx}$ corresponding to a single basin of attraction --
if we set $\vx^{0} \gets \vx$ and apply the update in \eqref{eq:att-dyn-finite} on $\vx^{0}$ for $T$ time-steps,
then $\vx^{T} \approx \brho_{\widehat\mu_\vx}$.
Furthermore, for appropriate $\beta$, $\widehat\mu_\vx$
matches the discrete assignment $\mu^\star_\vx$ in the $k$-means objective (\eqref{eq:kmeans-obj}).

This implies that, for appropriately set
$\beta$ and $T$, $\vx^{T} \approx \brho_{\mu^\star_\vx}$, allowing us to replace the desired per-example loss $\|\vx - \brho_{\mu^\star_\vx}\|^2$ in the $k$-means objective (\eqref{eq:kmeans-obj}) with $\|\vx - \vx^{T}\|^2$, allowing us to rewrite $k$-means (\eqref{eq:kmeans-obj}) as the following continuous optimization problem:
\begin{equation} \label{eq:kmeans-am-rel}
\min_\mR
\sum\nolimits_{\vx \in S}
\left\| \vx - \vx^{T}_\mR \right \|^2, \text{ where } \vx^{0} \gets \vx
\end{equation}
where $\vx^{T}_\mR$ is obtained by applying update in \eqref{eq:att-dyn-finite} to any $\vx \in S$ through $T$ recursion steps, and the subscript `${\cdot}_\mR$' highlights the dependence on the prototypes $\mR$.

By the above discussion, the objective in \eqref{eq:kmeans-am-rel} possesses the desired  {\em discrete assignment to a single prototype} of the original $k$-means objective (\eqref{eq:kmeans-obj}) since $\vx^{T}_\mR$ converges to exactly one of the prototypes. This is a significantly different relaxation of the $k$-means objective compared to the existing ``weighted-sum'' approaches.
The tightness of this relaxation
relies on the choices of $\beta$ (inverse temperature) and $T$ (the recursion depth) -- we treat them as hyperparameters and select them in a data-dependent manner.
For any given $\beta$ and $T$, we can minimize \eqref{eq:kmeans-am-rel} with SGD (and variants) via backpropagation through time.
\begin{figure}[!!!t]
\centering
\begin{subfigure}{0.3\textwidth}
\centering
{\footnotesize
  \begin{tikzpicture}[scale=0.9]
  \filldraw[color=white!10, fill=Gray!5, very thin] (-0.1, -0.1) rectangle (5.1, 1.1);
  \fill[BurntOrange] (1, 0.5) circle (3pt) node[left] {$\brho_1\,$};
  \fill[ForestGreen] (4, 0.5) circle (3pt) node[right] {$\,\brho_2$};
  \fill[black] (2.1, 0.5) circle (2pt) node[below] {$\vx$};
  \draw[<-, BurntOrange, solid, thick] (1.2, 0.6) -- (2, 0.6); 
  \draw[->, ForestGreen, solid, thick] (2.2, 0.5) -- (2.8, 0.5);
  \draw[<-, black, solid, thick] (1.6, 0.4) -- (2, 0.4);
  \end{tikzpicture}%
}
\vspace{-0.1in}
\caption{Case I}
\label{fig:basin-1}
\end{subfigure}
~
\begin{subfigure}{0.3\textwidth}
\centering
{\footnotesize
  \begin{tikzpicture}[scale=0.9]
  \filldraw[color=white!10, fill=Gray!5, very thin] (-0.1, -0.1) rectangle (5.1, 1.1);
  \fill[BurntOrange] (1, 0.5) circle (3pt) node[left] {$\brho_1\,$};
  \fill[ForestGreen] (4, 0.5) circle (3pt) node (r2) [left] {$\brho_2\,$};
  \fill[NavyBlue] (4.5, 0.5) circle (3pt) node (r3) [right] {$\, \brho_3$};
  \fill[black] (2.1, 0.5) circle (2pt) node[below] {$\vx$};
  \draw[<-, BurntOrange, solid, thick] (1.2, 0.5) -- (2, 0.5); 
  \draw[->, ForestGreen, solid, thick] (2.2, 0.7) -- (2.9, 0.7);
  \draw[->, NavyBlue, solid, thick] (2.2, 0.5) -- (2.85, 0.5);
  \draw[->, black, solid, thick] (2.2, 0.3) -- (2.6, 0.3);
  \end{tikzpicture}
}\vspace{-0.1in}
\caption{Case II}
\label{fig:basin-2}
\end{subfigure}
~
\begin{subfigure}{0.3\textwidth}
\centering
{\footnotesize
  \begin{tikzpicture}[scale=0.9]
  \filldraw[color=white!10, fill=Gray!5, very thin] (-0.1, -0.1) rectangle (5.1, 1.1);
  \fill[BurntOrange] (1, 0.5) circle (3pt) node[left] {$\brho_1\,$};
  \fill[ForestGreen] (2.75, 0.9) circle (3pt) node[right] {$\,\brho_2$};
  \fill[NavyBlue] (2.75, 0.1) circle (3pt) node[right] {$\,\brho_3$};
  \fill[black] (2.1, 0.5) circle (2pt) node[below] {$\vx$};
  \draw[<-, BurntOrange, solid, thick] (1.45, 0.6) -- (2, 0.6); 
  \draw[->, ForestGreen, solid, thick] (2.2, 0.57) -- (2.65, 0.8);
  \draw[->, NavyBlue, solid, thick] (2.2, 0.43) -- (2.65, 0.2);
  \draw[<-, black, solid, thick] (1.6, 0.4) -- (2, 0.4);
  \end{tikzpicture}%
}
\vspace{-0.1in}
\caption{Case III}
\label{fig:basin-3}
\end{subfigure}
\vspace{-0.1in}
\caption{{\bf Collective attraction.} Examples of prototype configurations, with the attraction (direction \& magnitude) of  $\vx$ towards each prototype shown by colored arrows, and the aggregate attraction shown by the black arrow. See \S\ref{sec:clam:partitions}.%
\vspace{-0.2in}}
\label{fig:basins-examples}
\end{figure}
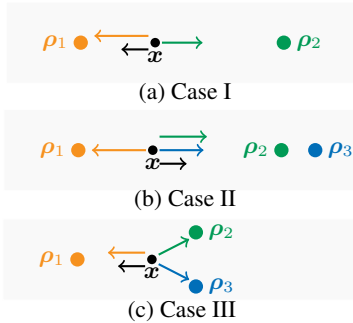
\subsection{Understanding the partitions induced by AMs} \label{sec:clam:partitions}
An equivalent interpretation of the discrete assignment $\mu^\star_\vx$ for each $\vx \in S$ in the $k$-means objective (\eqref{eq:kmeans-obj}) is that the prototypes $\mR$ induce a Voronoi partition~\citep{aurenhammer1991voronoi} of the whole input domain, and the examples $\vx \in S$ are assigned to the prototype whose partition they lie in. Voronoi partitions have piecewise linear boundaries, with all points in a partition having the same closest prototype.
Given prototypes $\mR$, the AM dynamics induces a partition  of the domain into non-overlapping basins of attraction.
For an appropriately set $\beta$ and $T$, the basins of attraction approximately match the Voronoi partition induced by the prototypes. However, when $\beta$ is not ``large enough'', the cluster assignment happens in a more ``collective'' manner.
We provide some intuitive examples here.

Consider two prototypes $\brho_1, \brho_2$, with a point $\vx$  between them as shown in \figref{fig:basin-1}, with $\brho_1$ closest to $\vx$. The attraction of $\vx$ to $\brho_1$ is inversely proportional to $\|\vx - \brho_1\|^2$ (vice versa for $\brho_2$). In this scenario, these will work against each other and $\vx$ will {\em move toward} the prototype with the highest attraction (based on \eqref{eq:att-dyn-finite}), which is the prototype closest to $\vx$ (in this case, $\brho_1$). This further increases the attraction of $\vx$ to $\brho_1$ in the next time-step (and decreases the $\brho_2$ attraction), resulting in $\vx$ progressively converging to $\brho_1$ through the $T$ time-steps, and thus being ``assigned'' to the partition corresponding to $\brho_1$, its closest prototype.

Consider an alternate Case-II with three prototypes in \figref{fig:basin-2}, with $\brho_2$ lying between $\brho_1$ and $\brho_3$, and $\vx$ lying between $\brho_1$ and $\brho_2$, though closer to $\brho_1$. For a large enough $\beta$, $\vx$ will converge to $\brho_1$ through the $T$ recursions. However, if $\beta$ is not large enough, the collective attraction of $\vx$ to $\brho_2$ and $\brho_3$ can overcome the largest single attraction of $\vx$ to $\brho_1$, forcing $\vx$ to move away from $\brho_1$, and eventually converge to $\brho_2$ -- while the dynamics guarantee convergence to exactly one of the prototypes, the collective effort of $\brho_2$ and $\brho_3$ causes $\vx$ to {\bf not converge to its closest prototype}.

While \figref{fig:basin-2} provides an example where prototypes can combine their attractions, \figref{fig:basin-3} presents Case-III  with three prototypes where prototypes {\bf cancel} their attraction. Here, prototypes $\brho_2$ and $\brho_3$ are closer to the example $\vx$ than $\brho_1$. However, for some appropriate value of $\beta$, the attraction of $\vx$ to $\brho_2$ and $\brho_3$ will cancel each other, allowing the attraction of $\vx$ to $\brho_1$ to move the example $\vx$ towards and finally converge to $\brho_1$ (which is the farthest of the three).
\begin{figure}[t]
\centering
\begin{subfigure}{0.27\columnwidth}
\centering
\centerline{\includegraphics[width=\columnwidth]{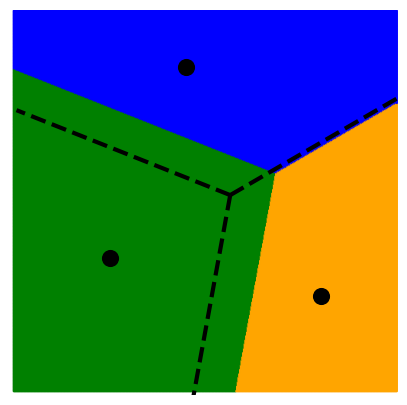}}
\caption{$\beta$ = 0.001.}
\label{fig:beta_0.001}
\end{subfigure}
~
\begin{subfigure}{0.27\columnwidth}
\centering
\centerline{\includegraphics[width=\columnwidth]{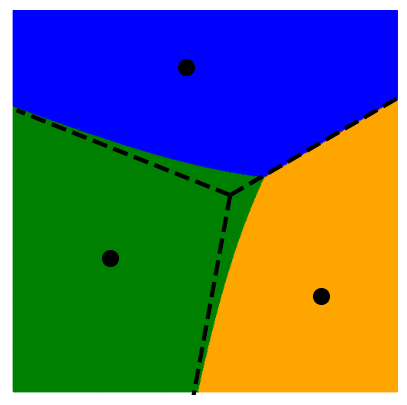}}
\caption{$\beta$ = 10.}
\label{fig:beta_10}
\end{subfigure}
~
\begin{subfigure}{0.27\columnwidth}
\centering
\centerline{\includegraphics[width=\columnwidth]{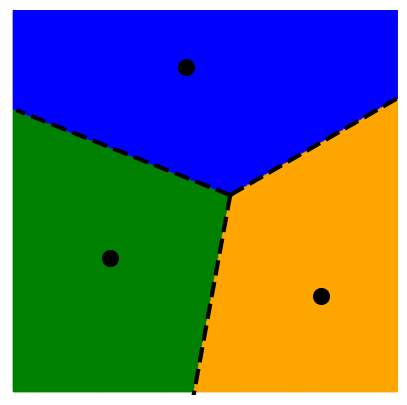}}
\caption{$\beta$ = 100.}
\label{fig:beta_100}
\end{subfigure}
\vspace{-0.1in}
\caption{{\bf Basins of attraction vs Voronoi partition.} Partitions induced by AM basins of attraction with given memories (black dots) for different $\beta$ are shown by the colored regions ($T$=10). Dashed lines show the Voronoi partition.
\vspace{-0.2in}}
\label{fig:am-vs-voronoi}
\end{figure}

These examples highlight that, for some values of $\beta$, the partition is a collective operation (involving all prototypes) leading to non-Voronoi partitions. We visualize the basins of attraction for different $\beta$ in \figref{fig:am-vs-voronoi} (also \S \ref{asec:add-exp:clam-voronoi}, \figref{fig:asec:add-exp:am-vs-voronoi}), contrasting them to the Voronoi partition.
For a small $\beta$ (\figref{fig:beta_0.001}), the basins do not match the Voronoi partition -- this mirrors the aforementioned behavior in Case-III (\figref{fig:basin-3}). As $\beta$ increases, the basins of attraction evolve to match the Voronoi partition (\figref{fig:beta_100}).
Furthermore, while the Voronoi partition boundaries are piecewise linear, the AM basins of attraction can have nonlinear boundaries (see  green-orange and green-blue boundaries in \figref{fig:beta_10}).

This behavior is an artifact of our novel continuous relaxation of discrete clustering. While we highlight this as a difference (ultimately, we are relaxing a discrete problem into a continuous one), this behavior requires specific values of $\beta$ relative to the specific geometric positioning of the prototypes; this might not be as prevalent in multidimensional data. Moreover, even if the basins do not match the Voronoi partition, it will not affect the clustering objective unless some $\vx \in S$ lies in the space where the partitions differ. Finally, as mentioned earlier, we treat $\beta$ and $T$ as hyperparameters, and select them in a data-dependent fashion, so that the AM partition sufficiently matches the Voronoi one.
\subsection{\clam: Clustering with AMs and self-supervision}\label{sec:clam:alg}
The AM framework allows for a novel end-to-end differentiable unconstrained continuous relaxation (\eqref{eq:kmeans-am-rel}) of the discrete $k$-means problem (\eqref{eq:kmeans-obj}).
Next, we wish to leverage the aforementioned strong pattern completion abilities of AMs (\S \ref{sec:clam:am}). We use standard masking from self-supervised learning
-- we apply a (random) mask $\vm \in \{0, 1\}^d$ to a point $\vx \in S \subset \R^d$ and utilize the AM recursion to complete the partial pattern $\vx^{0} = \vm \odot \vx$ and utilize the distortion between the ground-truth masked value and completed pattern as our loss as follows. Given a mask distribution $\gM$ and denoting $\bar \vm$ as the complement of $\vm$:
\begin{equation} \label{eq:clam:masked}
\mathcal L =  
\sum\nolimits_{\vx \in S} \E_{\vm \sim \gM}
  \|
  \bar \vm \odot (\vx - \vx^{T}_\mR ) 
  \|^2,
  \, \vx^{0} \gets \vm \odot \vx,
\end{equation}
where $\vx^{T}_\mR$ evolves from $\vx^{0}$ with $T$ recursions of \eqref{eq:att-dyn-finite}. Prototypes  $\mR$ are learned by minimizing the self-supervised pattern completion loss (\eqref{eq:clam:masked}) via SGD.
\begin{figure}[t]
\begin{center}
{\tiny
\begin{tikzpicture}[scale=0.2]
  \newcommand \bx {6}
  \newcommand \by {3}
  \draw[color=Gray, thin, fill=Gray!10] (\bx+1.5, \by+0.5) rectangle (\bx+25.75, \by+20.5);
  \node (R) at (\bx+20.5, \by+1.5) {$\mR = \{ \brho_1, \brho_2, \brho_3 \}$};
  \foreach \i in {1,...,5}
  {
    \filldraw[color=BurntOrange, fill=BurntOrange!60, thick] (\bx+3, \by+2+\i) rectangle (\bx+4, \by+3+\i);
  }
  \node (r1) at (\bx+2.5, \by+9) {$\brho_1$};
  \foreach \i in {1,...,5}
  {
    \filldraw[color=ForestGreen, fill=ForestGreen!60, thick] (\bx+11, \by+0+\i) rectangle (\bx+12, \by+1+\i);
  }
  \node (r2) at (\bx+10, \by+2) {$\brho_2$};
  \foreach \i in {1,...,5}
  {
    \filldraw[color=NavyBlue, fill=NavyBlue!60, thick] (\bx+24, \by+4+\i) rectangle (\bx+25, \by+5+\i);
  }
  \node (r3) at (\bx+23, \by+6) {$\brho_3$};
  \foreach \i in {1,...,5}
  {
    \filldraw[color=black, fill=white, thick] (\bx+10, \by+13+\i) rectangle (\bx+11, \by+14+\i);
  }
  \foreach \i in {1,...,2}
  {
    \filldraw[color=brown, fill=Plum, thick] (\bx+10, \by+13+\i) rectangle (\bx+11, \by+14+\i);
  }
  \node (vv) at (\bx+16, \by+19) {$\vx^{0} \gets \vm \odot \vx$};
  \node (vv1) at (\bx+10.5, \by+11) {$\vx^{1}_\mR$};
  \draw[very thick, dotted, ->, black] (\bx+9.75, \by+14.5) -- (\bx+8.75, \by+11);
  \foreach \i in {1,...,2}
  {
    \filldraw[color=black, fill=BurntOrange!60, thick] (\bx+3.5, \by+2.5+\i) rectangle (\bx+4.5, \by+3.5+\i);
  }
  \foreach \i in {3,...,5}
  {
    \filldraw[color=black, fill=white, thick] (\bx+3.5, \by+2.5+\i) rectangle (\bx+4.5, \by+3.5+\i);
  }
  \node (vvT) at (\bx+6.5, \by+4.5) {$\vx^{T}_\mR$};
  \node (vv2) at (\bx+9.5, \by+9) {$\vx^{2}_\mR$};
  \draw[very thick, dotted, ->, black] (\bx+8.75, \by+11) -- (\bx+7.75, \by+9);
  \node (vvt) at (\bx+8.5, \by+7) {$\vx^{t}_\mR$};
  \draw[very thick, dotted, ->, black] (\bx+7.75, \by+9) -- (\bx+6.75, \by+7);
  \draw[very thick, dotted, ->, black] (\bx+6.75, \by+7) -- (\bx+5.5, \by+5.5);

  \foreach \i in {1,...,5}
  {
    \filldraw[color=black, fill=black!30, thick] (2, \by+2+\i) rectangle (3, \by+3+\i);
  }
  \node (x) at (2.5, \by+2) {$\vx$};
  \foreach \i in {1,...,5}
  {
    \filldraw[color=black, fill=black!30, thick] (2, \by+13+\i) rectangle (3, \by+14+\i);
  }
  \filldraw[color=brown, fill=Plum, thick] (3.5, \by+14) rectangle (4.5, \by+15);
  \filldraw[color=brown, fill=Plum, thick] (3.5, \by+15) rectangle (4.5, \by+16);
  \node (m) at (4, \by+13) {$\vm$};
  \draw[very thick, ->, black] (2.5, \by+9) -- (2.5, \by+13);
  \draw[very thick, ->, black] (3.5, \by+16.5) -- (\bx+1, \by+16.5);
  \foreach \i in {1,...,5}
  {
    \filldraw[color=black, fill=black!30, thick] (\bx+30, \by+13+\i) rectangle (\bx+31, \by+14+\i);
  }
  \foreach \i in {1,...,2}
  {
    \filldraw[color=black, fill=BurntOrange!60, thick] (\bx+32, \by+13+\i) rectangle (\bx+33, \by+14+\i);
  }
  \foreach \i in {3,...,5}
  {
    \filldraw[color=black, fill=black!30, thick] (\bx+32, \by+13+\i) rectangle (\bx+33, \by+14+\i);
  }
  \foreach \i in {1,...,5}
  {
    \filldraw[color=BurntOrange, fill=BurntOrange!60, thick] (\bx+30, \by+4+\i) rectangle (\bx+31, \by+5+\i);
  }
  \foreach \i in {1,...,5}
  {
    \filldraw[color=ForestGreen, fill=ForestGreen!60, thick] (\bx+31.2, \by+4+\i) rectangle (\bx+32.2, \by+5+\i);
  }
  \foreach \i in {1,...,5}
  {
    \filldraw[color=NavyBlue, fill=NavyBlue!60, thick] (\bx+32.4, \by+4+\i) rectangle (\bx+33.4, \by+5+\i);
  }
  \draw[very thick, ->, black] (\bx+26.1, \by+16.5) -- (\bx+29.5, \by+16.5);
  \node (loss) at (\bx+29.3, \by+19) [align=left] {Self-supervised \\ loss $\mathcal{L}$};
  \draw[very thick, ->, black] (\bx+31.5, \by+13.5) -- (\bx+31.5, \by+10.5);
  \node (grad) at (\bx+29, \by+12) {$\nabla_\mR \mathcal{L}$};
  \draw[very thick, <-, black] (\bx+26.1, \by+7.5) -- (\bx+29.5, \by+7.5);
  \node (um) at (\bx+29.5, \by+3.7) [align=left] {Updated \\ prototypes $\mR$};
  \filldraw[color=Gray, fill=Gray!50, thick] (\bx+13, \by+9) rectangle (\bx+20, \by+11);
  \node (amr) at (\bx+16.5, \by+10) [align=left] {\bf AM recursion};
  \draw [->, very thick] (\bx+16.5, \by+11.25) arc [start angle=35, end angle=335, x radius=2.5, y radius=2.5];
\end{tikzpicture}
}
\end{center}
\vspace{-0.1in}
\caption{{\bf \Algref{alg:clam}.} For $\vx \in S$, we first apply a mask (in \textcolor{Plum}{purple}) $\vm$ to $\vx$ to get the initial iterate $\vx^{0}$ for the AM recursion. With $T$ recursions, we have a completed version $\vx^{T}_\mR$. The prototypes $\mR$ are updated with the gradient $\nabla_\mR \mathcal{L}$ on the self-supervised  loss $\mathcal{L}$ (\eqref{eq:clam:masked}).%
\vspace{-0.2in}}
\label{fig:clam-viz}
\end{figure}
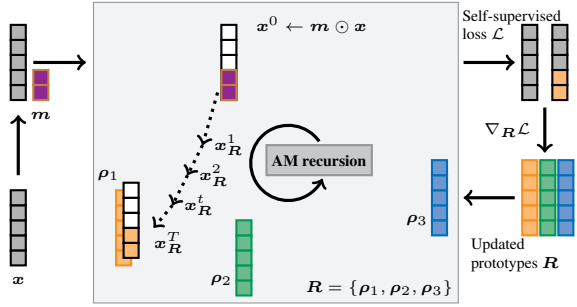
%
%
\begin{algorithm}[t]
\caption{\clam: Learning $k$ prototypes for clustering a $d$-dimensional dataset $S \subset \R^d$ into $k$ clusters with $T$ time-steps for $N$ epochs, with inverse temperature $\beta$, learning rate $\eps$, time step $dt$ and time constant $\tau$.}
\label{alg:clam}
\DontPrintSemicolon
{\footnotesize
\SetKwProg{train}{Train{\clam}$(S, k, N, T, \beta, dt, \tau, \eps)$}{}{end}
\SetKwProg{infer}{Infer{\clam}$(S, \mR, T, \beta, dt, \tau)$}{}{end}
\train{}{
Initialize prototypes $\mR = \{ \brho_\mu \in \R^d, \mu \in [k] \}$ randomly\;
\For{epoch $n = 1, \ldots, N$}{
  \For{batch $B \in S$}{
    Batch loss $\mathcal L_B  \gets 0$\;
    \For{example $\vx \in B$}{
      Sample random mask $\vm \in \{0, 1\}^d$\;
      $\vx^{0} \gets \vm \odot \vx$\;
      \For{$t = 1, \ldots, T$}{
        $\Delta_\mu \gets \brho_\mu - \vx^{t-1} _\mR \, \;\;\forall \mu \in [k]$\;
        $\bdelta^{t} \gets \bar\vm \odot \frac{dt}{\tau} \sum\limits_{\mu=1}^k \Delta_\mu \softmax \left(-\beta \| \Delta_\mu \|^2 \right)$\;
        $ \vx^{t}_\mR \gets \vx^{t-1}_\mR + \bdelta^{t}$\;
      }
      $\mathcal L_B \gets \mathcal L_B + \|\bar \vm \odot ( \vx^{T}_\mR -  \vx ) \|^2$\;

    }
    $\brho_\mu \gets \brho_\mu - \eps \nabla_{\brho_\mu} \mathcal L_B\, \;\;\forall \mu = 1, \ldots, k$
  }
}
\KwRet{Protoypes $\mR = \{\brho_\mu, \mu \in [k] \}$}
}
\infer{}{
Cluster assignments $C \gets \emptyset$\;
\For{$\vx \in S$}{
  $\vx^{0} \gets \vx$\;
  \For{$t = 1, \ldots, T$}{
    $\Delta_\mu \gets \brho_\mu - \vx^{t-1}_\mR \, \;\;\forall \mu \in [k]$\;
    $ \vx^{t}_\mR \gets \vx^{t-1}_\mR + \frac{dt}{\tau} \sum\limits_{\mu=1}^k \Delta_\mu \softmax \left(-\beta \| \Delta_\mu \|^2 \right)$\;
  }
  $\widehat\mu_\vx \gets \argmin_\mu \| \brho_\mu - \vx^{T}_\mR \|^2$\;
  $C \gets C \cup \{\widehat\mu_\vx\}$\;
}
\KwRet{Per-point cluster assignments $C$}
}
}
\end{algorithm}
%
%
The precise learning algorithm is presented in the Train{\clam} subroutine of \algref{alg:clam}, and visualized in \figref{fig:clam-viz}. After randomly seeding the prototypes (line 2), we perform $N$ epochs over the data (line 3) -- to each $\vx$ in a data batch $B$ (lines 4-6), we apply a random mask (line 8), complete the pattern with the AM recursion using the current prototypes (lines 9-12), accumulate the batch loss as in \eqref{eq:clam:masked} (line 13), and update the prototypes with the batch gradient (line 14).
\begin{proposition} \label{prop:train-time}
The Train{\clam} subroutine in \algref{alg:clam} takes $O(dkTN |S|)$ time, where $|S|$ is the cardinality of $S$,  converging to a $O(N^{-1/2})$-stationary point.
\end{proposition}
The $N^{-1/2}$ term comes from the convergence rate of standard SGD for smooth non-convex problems~\citep{nesterov2003introductory} that can be improved to $N^{-2/3}$ with momentum~\citep{fang2018spider}, which is straightforward given the end-to-end differentiability of our proposal.
The Infer{\clam} subroutine of \algref{alg:clam} assigns points $\vx \in S$ to clusters using the learned prototypes $\mR$ with a single pass over $S$.
This is invoked once at the conclusion of the clustering.
%
\begin{proposition} \label{prop:infer-time}
The Infer{\clam} subroutine in \algref{alg:clam} takes $O(dkT |S|)$ time, where $|S|$ is the cardinality of $S$.
\end{proposition}

From Propositions \ref{prop:train-time} and \ref{prop:infer-time}, we can see, \clam training and inference is linear in the number of points $|S|$ in the data set $S$, the number of dimensions $d$, and the number of clusters $k$. For $N$ epochs over the data set $S$, \clam training takes $O(dkTN |S|)$ where $T$ is the AM recursion depth, while $k$-means training (with Lloyd’s algorithm) takes $O(dkN |S|)$. Hence, both $k$-means and \clam scale linearly with the number of points, though \clam runtime has a multiplicative factor of $T$ where $T$ is usually less than 10. So $k$-means would intuitively be faster than \clam. Note that $k$-means and \clam are significantly more efficient than Spectral clustering and Agglomerative clustering which usually scale between quadratic and cubic in the number of samples. Note that, for the same number of epochs, we can establish convergence guarantees for our SGD based \clam (which can be improved with the use of Momentum based SGD), while similar convergence guarantees are not available for $k$-means.

\subsection{Flexibility and Extensions} \label{sec:clam:flex}
Here, we highlight how \clam naturally incorporates modifications. Specifically, we demonstrate how it can be modified to perform weighted clustering (different clusters have different ``attractions''), and spherical clustering (using  cosine similarity), highlighting the change in the energy function (\eqref{eq:particle-energy}) and the attractor dynamics (\eqref{eq:att-dyn-finite}).

\paragraph{Weighted clustering.}
We allow different prototypes $\brho_\mu$ to have different weights $\varepsilon_\mu > 0$, implying different attraction scaling, and the subsequent energy function is:
\begin{equation}\label{eq:energy:weights}
  E(\vv) = -\frac{1}{\beta}\log \left(\sum_{\substack{\mu\in [k]}} \varepsilon_\mu  \exp \left( -\beta \| \brho_\mu - \vv \|^2 \right)\right)
\end{equation}
with the finite difference update $\bdelta^{t+1}$ at time-step $t+1$
\begin{equation}\label{eq:att-dyn-finite:weights}
\bdelta^{t+1} = 
\frac{dt}{\tau} \sum_{\mu\in[k]} (\brho_\mu - \vv^{t} ) \softmax (-\beta \| \brho_\mu - \vv^{t} \|^2 + \log \varepsilon_\mu )
\end{equation}
This update step can be directly plugged into \algref{alg:clam} (line 11). The weights $\varepsilon_\mu, \mu \in [k]$ can be user-specified based on domain knowledge, or learned via gradient descent, and intuitively relate to the slopes of the basins of attraction, allowing different attraction scaling for each basin.

\paragraph{Spherical clustering.}
We can change the energy function to depend on the cosine similarity to get (without loss of generality, assume $\|\vv\| = 1\,\; \forall \vv \in S$ in the following):
\begin{equation}\label{eq:energy:spherical}
  E(\vv) = -\frac{1}{\beta}\log \left(\sum\nolimits_{\mu\in[k]}  \exp \left( \beta \left\langle \tilde{\brho_\mu}, \vv \right\rangle \right)\right)
\end{equation}
with $\tilde\brho_\mu = \nicefrac{\brho_\mu}{\| \brho_\mu \|}$, and the corresponding update step
\begin{equation}\label{eq:att-dyn-finite:spherical}\nonumber
\bdelta^{t+1} = \frac{dt}{\tau} \sum\nolimits_{\mu\in[k]} \tilde\brho_\mu  \softmax \left(\beta \left\langle \tilde\brho_\mu, \vv^{t} \right\rangle \right)
\tilde\vv = \vv^{t} + \bdelta^{t+1} 
\end{equation}
and $\vv^{t+1} = \nicefrac{\tilde\vv}{\| \tilde\vv \|}$,
plugged into \algref{alg:clam} (line 11) gives us an end-to-end differentiable spherical clustering.

\section{Empirical evaluation} \label{sec:eval}
We evaluate \clam on 10 datasets of varying sizes -- 7-16000 features, 101-60000 points. The number of clusters for each dataset are selected based on its number of underlying classes  (the class information is not involved in clustering or hyperparameter selection). See details in Appendix~\ref{asec:exp-details:data}. First, we compare \clam with established schemes, $k$-means~\citep{lloyd1982least}, spectral \& agglomerative clustering, and to DEC~\citep{xie2016unsupervised} and DCEC~\cite{guo2017deep} (for 3 image sets since DCEC uses CAE  (\S \ref{sec:related_work}); \clam naively reshapes images into vectors).
Then we ablate the effect of self-supervision in \clam, and evaluate a \clam extension. 
Qualitatively, we visualize the memory evolution through the learning.

For implementation, we use Tensorflow~\citep{abadi2016tensorflow} for \clam and {\tt scikit-learn}~\citep{pedregosa2011scikit} for the clustering baselines and quality metrics.
DCEC/DEC  seed the optimization with the prototypes from the Lloyd's solution; we also consider a randomly seeded DEC$^r$. 
Further details are in Appendix~\ref{asec:expt-details:implementation}. We perform an elaborate hyperparameter search for all methods (see Appendices \ref{asec:add-exp:clam-hpo} and \ref{asec:expt-details:hpo}), and utilize the configuration for each method corresponding to the best Silhouette Coefficient (SC)~\citep{rousseeuw1987silhouettes} on each dataset. The code is available at \url{https://github.com/bsaha205/clam}.

\begin{table*}[t]
\caption{{\bf Silhoutte Coefficient (SC) obtained by \clam, its variants and baselines ({\em higher is better}).} \up (\down) show the performance gain (drop) by \clam, given by $\nicefrac{(a - b)}{b}\times 100\%$ where $a$ is the SC for \clam, and $b$ is the SC for the baseline; positive (negative) values indicate performance gain (drop). The best performance for a dataset is in {\bf boldface}. W/L denote Wins/Losses. Note ablations \clam w/ (\ref{eq:kmeans-am-rel}) is \clam without self-supervision, and \clam w/ (\ref{eq:att-dyn-finite:weights}) is weighted \clam.%
\vspace{-0.15in}}
\label{tab:ss}
\begin{center}
{\footnotesize
\begin{tabular}{lllllll|lll}
\toprule
\multicolumn{1}{l}{\bf Dataset}  &\multicolumn{1}{l}{\bf $k$-means} &\multicolumn{1}{l}{\bf Spectral} &\multicolumn{1}{l}{\bf Agglo}  &\multicolumn{1}{l}{\bf DCEC}  &\multicolumn{1}{l}{\bf DEC} &\multicolumn{1}{l}{\bf DEC$^r$} & \multicolumn{1}{|l}{\bf \clam} & {\bf w/ (\ref{eq:kmeans-am-rel})} & {\bf w/ (\ref{eq:att-dyn-finite:weights})}\\
\midrule
Zoo & 0.374\txsps{\up 10\%} & 0.398\txsps{\up 3\%} & 0.398\txsps{\up 3\%} & --- & 0.374\txsps{\up 10\%} & 0.367\txsps{\up 12\%} & \textbf{0.412} & 0.382 & \textbf{0.412} \\
Ecoli & 0.262\txsps{\up 26\%} & 0.381\txsps{\down 13\%} & \textbf{0.404}\txsps{\down 18\%} & --- & 0.262\txsps{\up 26\%} & 0.255\txsps{\up 30\%} & 0.331 & 0.301 & {\em 0.345} \\
MovLib &
0.252\txsps{\up 3\%} & 0.247\txsps{\up 5\%} & 0.247\txsps{\up 5\%} & --- & 0.258\txsps{\up 1\%} & 0.184\txsps{\up 41\%} & \textbf{0.260} & 0.248 & {\em 0.262} \\
Yale & 0.117\txsps{\up 10\%} & 0.120\txsps{\up 8\%} & 0.114\txsps{\up 13\%} & 0.089\txsps{\up 45\%} & 0.123\txsps{\up 5\%} & 0.088\txsps{\up 47\%} & \textbf{0.129} & 0.121  & 0.119  \\
USPS & 0.135\txsps{\up 1\%} & 0.228\txsps{\down 40\%} &  \textbf{0.231}\txsps{\down 41\%} & 0.119\txsps{\up 14\%} & 0.135\txsps{\up 1\%} & 0.108\txsps{\up 26\%} & 0.136  & 0.127 & 0.136 \\
Segment & 0.357\txsps{\up 35\%} &\textbf{0.702}\txsps{\down 31\%} & 0.697\txsps{\down 31\%} & ---  &  0.357\txsps{\up 35\%} &  0.287\txsps{\up 68\%} & 0.483 & 0.357 & 0.410\\
FMNIST &
0.126\txsps{\up 10\%} & 0.075\txsps{\up 84\%} &  0.096\txsps{\up 44\%} & 0.121\txsps{\up 14\%} & 0.129\txsps{\up 7\%} & 0.070\txsps{\up 97\%} & \textbf{0.138} & 0.131 & 0.118\\ 
GCM & 0.118\txsps{\up 62\%} & 0.205\txsps{\down 7\%} & \textbf{0.288}\txsps{\down 34\%} & --- & 0.115\txsps{\up 66\%} & 0.054\txsps{\up 254\%} & 0.191 & 0.142 & 0.156 \\
MicePE & 0.128\txsps{\up 56\%} & 0.156\txsps{\up 28\%} & 0.193\txsps{\up 4\%} & --- & 0.137\txsps{\up 46\%} & 0.115\txsps{\up 74\%} & \textbf{0.201} & 0.126 & 0.127 \\
CTG & 0.161\txsps{\up 53\%} &\textbf{0.449}\txsps{\down 45\%} & 0.387\txsps{\down 36\%} & --- & 0.164\txsps{\up 50\%} & 0.130\txsps{\up 89\%} & 0.246 & 0.147 & 0.158 \\
\midrule
\clam W/L & 10/0 & 5/5 & 5/5 &  3/0 & 10/0 & 10/0 & --- & 10/0 & 6/2 \\
\clam w/ (\ref{eq:kmeans-am-rel}) W/L & 5/3 & 3/7 & 3/7 &  3/0 & 5/4 & 10/0 & 0/10 & --- & 2/8 \\
\bottomrule

\end{tabular}
}
\end{center}
\vspace{-0.25in}
\end{table*}
\begin{table}[t]
\caption{{\bf Normalized Mutual Information (NMI) between ground-truth labels and clusters ({\em higher is better}).}
The best performance for each dataset is in {\bf boldface}. The \underline{underlined} entries for {\bf Spec}tral and {\bf Aggl}omerative mark low NMI but high SC in Table~\ref{tab:ss} (note abbreviations $k$-means$\to${$k$}-m, DCEC$\to$DC, DEC$\to$D, DEC$^r${$\to$}D$^r$).%
\vspace{-0.15in}}
\label{tab:nmigt}
\begin{center}
{\footnotesize
\begin{tabular}{lccccccc}
\toprule
\multicolumn{1}{l}{\bf Data}  &\multicolumn{1}{c}{\bf $k$-m} &\multicolumn{1}{c}{\bf Spec} &\multicolumn{1}{c}{\bf Aggl} &\multicolumn{1}{c}{\bf DC} & \multicolumn{1}{c}{\bf D} & \multicolumn{1}{c}{\bf D$^r$} &\multicolumn{1}{c}{\bf \clam}\\
\midrule
Zoo & 0.83 & 0.89 & 0.84 & N/A & 0.83 & 0.80 & \textbf{0.94} \\
Yale & 0.60 & 0.57 & \textbf{0.67} & 0.54 & 0.55 & 0.51 & 0.64 \\
GCM & 0.44 & \underline{0.17} & \underline{0.19} & N/A & 0.42 & 0.39 & \textbf{0.45} \\
Ecoli & 0.63 & 0.66 & \textbf{0.71} & N/A & 0.63 & 0.57 & 0.66 \\
MLib &
0.60 & 0.61 & 0.61 & N/A & 0.60 & 0.51 & \textbf{0.62} \\
MPE &
0.24 & 0.01 & 0.26 & N/A & 0.29 & 0.30 & \textbf{0.31} \\
USPS & 0.54 & \underline{0.08} & \underline{0.02} & \textbf{0.69} & 0.54 & 0.45 & 0.56 \\
CTG & \textbf{0.36} & \underline{0.04} & \underline{0.04} & N/A & 0.35 & \textbf{0.36} & 0.32 \\
Seg & 0.58 & \underline{0.01} & \underline{0.01} & N/A & 0.59 & \textbf{0.61} & 0.55 \\
FM & 0.50 & \textbf{0.64} & 0.01 & 0.59 & 0.50 & 0.43 & 0.52 \\
\bottomrule
\end{tabular}
}
\end{center}
\vspace{-0.25in}
\end{table}

\paragraph{Q1: How does \clam compare against baselines?}
We present the best SC obtained by all schemes in Table~\ref{tab:ss}.
\clam consistently improves over $k$-means across all 10 datasets (up to 60\%), and outperforms both versions of DEC, highlighting the advantage of the novel relaxation. Furthermore, Lloyd's seeding is critical in DEC -- DEC$^r$ does worse than base $k$-means (via \citet{lloyd1982least}) in all cases -- while {\em \clam performs well with random seeding}. \clam even improves upon DCEC, which uses rich image representations from a CAE.
Overall, \clam performs best on 5/10
datasets, showing significant improvements over even spectral and agglomerative clustering (these are not prototype-based, and hence can be more expressive). On the remaining datasets, both spectral and agglomerative clustering show improvements over \clam.
To understand this better, we also compare the clusters generated by the different methods to the ground-truth class structure in each of the datasets.%
\footnote{Although the underlying geometry of the data does not necessarily align with the ground-truth class structure, this is a {\em post-hoc way} of evaluating how intuitive the discovered clusters are.}
The Normalized Mutual Information (NMI) scores ~\citep{vinh2009information} are shown in Table~\ref{tab:nmigt}.
The results indicate that the datasets on which spectral and agglomerative clustering have a significantly higher SC than $k$-means and \clam (GCM, USPS, CTG, Segment), their corresponding NMI scores are significantly lower (see \underline{underlined} entries in Table~\ref{tab:nmigt}), indicating clusters that are misaligned with the ground-truth labels. Upon further investigation, we see that both spectral and agglomerative clustering end up with a single large cluster, and many really small (even singleton) ones, indicating that the clustering is overly influenced by geometric outliers.~\footnote{For example, with CTG (2126 points), both spectral and agglomerative clustering find 10 clusters, but 8 of them of size $\leq 8$ each, and 1 with size over 2000. $k$-means finds 1 cluster of size 7, with 8 of the remaining 9 clusters of size $\geq 91$. \clam finds 2 clusters with size $\leq 4$; all remaining have size $\geq 55$.}
See further results in Appendix~\ref{asec:add-exp:baseline}.

\paragraph{Q2: How beneficial is self-supervision in \clam?}
In \algref{alg:clam} (line 13), we use the pattern completion ability of AMs, and optimize for the self-supervised loss (\eqref{eq:clam:masked}). Here, we ablate the effect of this choice -- we utilize a version of \algref{alg:clam} that does not use any masking (removing $\vm$ and $\bar \vm$ in lines 8 and 11 respectively) and learns memories by minimizing the loss in \eqref{eq:kmeans-am-rel} (replacing line 13 with this loss).
Table~\ref{tab:ss} ({\bf \clam} and {\bf \clam w/ (\ref{eq:kmeans-am-rel})} columns) show their SC for all datasets,
highlighting the positive effect of self-supervision, with around 10\% gain in all cases.

\paragraph{Q3: How do weighted and vanilla clustering with \clam compare against each other?}
We discussed the flexibility of the \clam framework in \S \ref{sec:clam:flex}. We evaluate one such extension -- weighted clustering where different memories can have different relative attractions, inducing more imbalanced clustering which may be desirable in certain applications. \clam handles relative weights that are (i)~user-specified, or (ii)~learned via SGD within \algref{alg:clam} (with an additional update step for $\varepsilon_\mu$).
We consider case~(ii) here, learning $\varepsilon_\mu$ via SGD, and compare to vanilla \clam in Table \ref{tab:ss} ({\bf \clam} and {\bf \clam w/ (\ref{eq:att-dyn-finite:weights})} columns) on all datasets. The results indicate that the benefits are dataset-dependent. Our intent here is to show that \clam handles such problems if needed, and can result in improvements (as in 2/10 cases).

\begin{figure}[t]
\begin{center}
\begin{subfigure}{0.12\columnwidth}
\centering
    \includegraphics[width=0.8\textwidth]{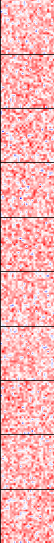}
  \caption{$n_0$}
  \label{fig:memory:e0}
\end{subfigure}
~
\begin{subfigure}{0.12\columnwidth}
  \includegraphics[width=0.8\textwidth]{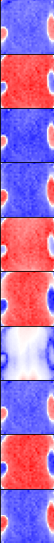}
  \caption{$n_5$}
  \label{fig:memory:e5}
\end{subfigure}
~
\begin{subfigure}{0.12\columnwidth}
  \includegraphics[width=0.8\textwidth]{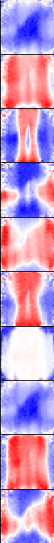}
  \caption{$n_{10}$}
  \label{fig:memory:e10}
\end{subfigure}
~
\begin{subfigure}{0.12\columnwidth}
  \includegraphics[width=0.8\textwidth]{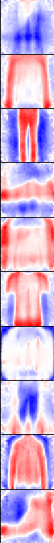}
  \caption{$n_{20}$}
  \label{fig:memory:e20}
\end{subfigure}
~
\begin{subfigure}{0.12\columnwidth}
  \includegraphics[width=0.8\textwidth]{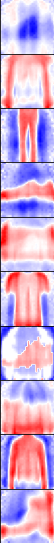}
  \caption{$n_{50}$}
  \label{fig:memory:e50}
\end{subfigure}
~
\begin{subfigure}{0.12\columnwidth}
\centering
  \includegraphics[width=0.8\textwidth]{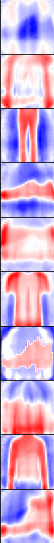}
  \caption{$n_{100}$}
  \label{fig:memory:e100}
\end{subfigure}
\end{center}
\vspace{-0.15in}
\caption{{\bf Evolution of \clam prototypes.} We visualize the prototypes at the $n^{\text{th}}$ training epoch for $n = 0, 5, 10, 20, 50, 100$ (with $T=10$). The images pixels are colored as red for positive, white for zero and blue for negative values, with their intensity denoting  magnitude.%
\vspace{-0.2in}}
\label{fig:memory}
\end{figure}
\paragraph{Q4: How to interpret \clam?}
An interesting aspect of AMs is the prototype-based representation of memories~\citep{krotov2016dense}. We study this for \clam with the Fashion-MNIST images. We use \clam to partition the 60k images of fashion items (shirts, trousers, shoes, bags) into 10 clusters, and visualize the evolution of the 10 memories in \figref{fig:memory} over the course of \algref{alg:clam}.
We reshape the 28$\times$28 images into vectors in $\R^{784}$, learn prototypes in $\R^{784}$ and reshape them back to 28$\times$28 solely for visualization.
Each sub-figure in \figref{fig:memory} corresponds to a particular epoch during the training, and shows all 10 memories stacked vertically.
We can see that at epoch 0 (\figref{fig:memory:e0}) the memories are initialized with random values.
At epoch 5 (\figref{fig:memory:e5}), the memories start to become less random, but only start showing some discernible patterns at epoch 10 (\figref{fig:memory:e10}) which get refined by epoch 20 (\figref{fig:memory:e20}) and further sharpen by epoch 50 (\figref{fig:memory:e50}). By epoch 100 (\figref{fig:memory:e100}), the shapes have stabilized although some learning might still be happening.
Some memories evolve into individual forms (such as rows 3--pants, 4--shoes, 8--bags, 9--long-sleeve shirts and 10--boots), while others evolve into mixtures of 2 forms (such as rows 1--pants and shoes, 6--pants and short-sleeve shirts, 7--shoes and long-sleeve shirts). The remaining evolve into less distinguishable forms though we can still visualize some shapes (like in rows 2--long-sleeve shirts and 5--shoes).
\begin{table}[t]
\caption{{\bf Clustering efficiency comparison between $k$-means and \clam.}}
\label{tab:efficiency-main}
\begin{center}
{\footnotesize
\begin{tabular}{lcc|cc}
\toprule
\multirow{2}{*}{\bf Dataset} & \multicolumn{2}{c}{\bf Silhoutte Coefficient (SC)} & \multicolumn{2}{c}{\bf Time in second}\\
\cmidrule{2-5} 
& \bf $k$-means & \bf \clam & \bf $k$-means & \bf \clam\\

\midrule
Zoo & 0.374 & 0.412 & 2 & 18\\
Ecoli  & 0.262 & 0.331 & 4 & 47 \\
MovLib  & 0.252 & 0.260 & 6 & 67 \\
Yale& 0.117 & 0.129 & 9 & 88 \\
CIFAR-10 & 0.022 & 0.031 & 2995 & 3419 \\
JV & 0.109 & 0.118 & 1973 & 3933 \\
\bottomrule
\end{tabular}
}
\end{center}
\end{table}

\paragraph{Q5: How computationally expensive is \clam compared to $k$-means?}
While comparing the actual runtimes of $k$-means and \clam, we would like to note that we make use of the standard $k$-means implementation from scikit-learn, which utilizes numpy and scipy linear algebra packages under the hood, while our \clam implementation makes use of the Tensorflow framework, leveraging their auto-grad capabilities. Furthermore, Tensorflow is designed to seamlessly utilize GPUs if available, while scikit-learn does not have such capabilities out of the box. With these differences, the actual runtimes are not directly comparable. This is the reason why we give importance to runtime complexities instead of actual runtimes. However, notwithstanding these caveats, we present the precise runtimes of $k$-means and \clam (the Spectral and Agglomerative clustering baselines are significantly slower, and DEC has the same scaling as $k$-means for the same number of epochs) for six datasets. In Table \ref{tab:efficiency-main}, we present the Silhouette Coefficient (SC) and Training time for $k$-means (with Lloyd’s algorithm) and \clam for all of these six datasets. In all cases, \clam improves over $k$-means in terms of the SC. For the smaller datasets, we can see that $k$-means is almost an order of magnitude faster than \clam. However, a lot of it can be attributed to the overhead of leveraging the Tensorflow framework and SGD, instead of just performing small matrix multiplications with Numpy. We also consider a relatively larger dataset CIFAR-10 where the images are encoded into 1920-dimensional vectors with a pre-trained DenseNet~\citep{huang2017densely} and a million row dataset, Japanese Vowels (JV), from \url{https://openml.org}. In these larger datasets, $k$-means is still faster than \clam (as we had discussed earlier based on the theoretical runtime complexities) but the increase of \clam’s runtime is only between 1-2$\times$.
\section{Limitations and Future Work} \label{sec:conc}
In this paper, we present a novel continuous relaxation of the discrete prototype-based clustering problem leveraging the AM dynamical system that allows us to solve the clustering problem with SGD based schemes. We show that this relaxation better matches the essence of the discrete clustering problem, and empirical results show that our \clam approach significantly improves over standard prototype-based clustering schemes, and existing continuous relaxations. We note that \clam is still a prototype-based clustering scheme, hence inherits the limitations of prototype-based clustering.

Given the end-to-end differentiable nature of \clam, we will extend it to clustering with kernel similarity functions and Mahanalobis distances, and to deep clustering where we also learn a latent space. We plan to explore new energy functions and update dynamics that enable spectral clustering. Finally, given \clam's flexibility, we want to automatically estimate the number of clusters on a per-dataset basis, much like \citet{pelleg2000x} and \citet{hamerly2003learning}.
\section*{Acknowledgements}
This work was supported by the Rensselaer-IBM AI Research Collaboration (\url{http://airc.rpi.edu}), a part of the IBM AI Horizons Network. The work was done during Bishwajit Saha's externship at MIT-IBM Watson AI Lab, IBM Research. 
%
%
\bibliography{amc}

\begin{thebibliography}{75}
\providecommand{\natexlab}[1]{#1}
\providecommand{\url}[1]{\texttt{#1}}
\expandafter\ifx\csname urlstyle\endcsname\relax
  \providecommand{\doi}[1]{doi: #1}\else
  \providecommand{\doi}{doi: \begingroup \urlstyle{rm}\Url}\fi

\bibitem[Abadi et~al.(2016)Abadi, Barham, Chen, Chen, Davis, Dean, Devin,
  Ghemawat, Irving, Isard, et~al.]{abadi2016tensorflow}
Abadi, M., Barham, P., Chen, J., Chen, Z., Davis, A., Dean, J., Devin, M.,
  Ghemawat, S., Irving, G., Isard, M., et~al.
\newblock $\{$TensorFlow$\}$: a system for $\{$Large-Scale$\}$ machine
  learning.
\newblock In \emph{12th USENIX symposium on operating systems design and
  implementation (OSDI 16)}, pp.\  265--283, 2016.

\bibitem[Aguilar et~al.(2001)Aguilar, Ruiz, Riquelme, and
  Gir{\'a}ldez]{aguilar2001snn}
Aguilar, J.~S., Ruiz, R., Riquelme, J.~C., and Gir{\'a}ldez, R.
\newblock {SNN}: A supervised clustering algorithm.
\newblock In \emph{International Conference on Industrial, Engineering and
  Other Applications of Applied Intelligent Systems}, pp.\  207--216. Springer,
  2001.

\bibitem[Amit et~al.(1985)Amit, Gutfreund, and Sompolinsky]{amit1985storing}
Amit, D.~J., Gutfreund, H., and Sompolinsky, H.
\newblock Storing infinite numbers of patterns in a spin-glass model of neural
  networks.
\newblock \emph{Physical Review Letters}, 55\penalty0 (14):\penalty0 1530,
  1985.

\bibitem[Aurenhammer(1991)]{aurenhammer1991voronoi}
Aurenhammer, F.
\newblock Voronoi diagrams -- a survey of a fundamental geometric data
  structure.
\newblock \emph{ACM Computing Surveys (CSUR)}, 23\penalty0 (3):\penalty0
  345--405, 1991.

\bibitem[Bezdek et~al.(1984)Bezdek, Ehrlich, and Full]{bezdek1984fcm}
Bezdek, J.~C., Ehrlich, R., and Full, W.
\newblock Fcm: The fuzzy c-means clustering algorithm.
\newblock \emph{Computers \& geosciences}, 10\penalty0 (2-3):\penalty0
  191--203, 1984.

\bibitem[Bijuraj(2013)]{bijuraj2013clustering}
Bijuraj, L.
\newblock Clustering and its applications.
\newblock In \emph{Proceedings of National Conference on New Horizons in
  IT-NCNHIT}, volume 169, pp.\  172, 2013.

\bibitem[Burns \& Fukai(2023)Burns and Fukai]{burns2023simplicial}
Burns, T.~F. and Fukai, T.
\newblock Simplicial hopfield networks.
\newblock In \emph{The Eleventh International Conference on Learning
  Representations}, 2023.

\bibitem[Cai et~al.(2022)Cai, Wang, Xu, and Guo]{cai2022unsupervised}
Cai, J., Wang, S., Xu, C., and Guo, W.
\newblock Unsupervised deep clustering via contractive feature representation
  and focal loss.
\newblock \emph{Pattern Recognition}, 123:\penalty0 108386, 2022.

\bibitem[Chakraborty et~al.(2021)Chakraborty, Paul, and
  Das]{chakraborty2021automated}
Chakraborty, S., Paul, D., and Das, S.
\newblock Automated clustering of high-dimensional data with a feature weighted
  mean shift algorithm.
\newblock In \emph{Proceedings of the AAAI Conference on Artificial
  Intelligence}, volume~35, pp.\  6930--6938, 2021.

\bibitem[Chazan et~al.(2019)Chazan, Gannot, and Goldberger]{chazan2019deep}
Chazan, S.~E., Gannot, S., and Goldberger, J.
\newblock Deep clustering based on a mixture of autoencoders.
\newblock In \emph{2019 IEEE 29th International Workshop on Machine Learning
  for Signal Processing (MLSP)}, pp.\  1--6. IEEE, 2019.

\bibitem[Curtin et~al.(2013)Curtin, March, Ram, Anderson, Gray, and
  Isbell]{curtin2013tree}
Curtin, R., March, W., Ram, P., Anderson, D., Gray, A., and Isbell, C.
\newblock Tree-independent dual-tree algorithms.
\newblock In \emph{International Conference on Machine Learning}, pp.\
  1435--1443. PMLR, 2013.

\bibitem[Curtin et~al.(2015)Curtin, Lee, March, and Ram]{curtin2015plug}
Curtin, R.~R., Lee, D., March, W.~B., and Ram, P.
\newblock Plug-and-play dual-tree algorithm runtime analysis.
\newblock \emph{Journal of Machine Learning Research}, 16:\penalty0 3269--3297,
  2015.

\bibitem[Dasgupta(2008)]{dasgupta2008hardness}
Dasgupta, S.
\newblock \emph{The hardness of k-means clustering}.
\newblock Department of Computer Science and Engineering, University of
  California~…, 2008.

\bibitem[Demircigil et~al.(2017)Demircigil, Heusel, L{\"o}we, Upgang, and
  Vermet]{demircigil2017model}
Demircigil, M., Heusel, J., L{\"o}we, M., Upgang, S., and Vermet, F.
\newblock On a model of associative memory with huge storage capacity.
\newblock \emph{Journal of Statistical Physics}, 168\penalty0 (2):\penalty0
  288--299, 2017.

\bibitem[Dempster et~al.(1977)Dempster, Laird, and Rubin]{dempster1977maximum}
Dempster, A.~P., Laird, N.~M., and Rubin, D.~B.
\newblock Maximum likelihood from incomplete data via the em algorithm.
\newblock \emph{Journal of the Royal Statistical Society: Series B
  (Methodological)}, 39\penalty0 (1):\penalty0 1--22, 1977.

\bibitem[Donath \& Hoffman(1973)Donath and Hoffman]{donath1973lower}
Donath, W.~E. and Hoffman, A.~J.
\newblock Lower bounds for the partitioning of graphs.
\newblock \emph{IBM Journal of Research and Development}, 17\penalty0
  (5):\penalty0 420--425, 1973.
\newblock \doi{10.1147/rd.175.0420}.

\bibitem[Dua et~al.(2017)Dua, Graff, et~al.]{dua2017uci}
Dua, D., Graff, C., et~al.
\newblock Uci machine learning repository.
\newblock 2017.

\bibitem[Duchi et~al.(2011)Duchi, Hazan, and Singer]{duchi2011adaptive}
Duchi, J., Hazan, E., and Singer, Y.
\newblock Adaptive subgradient methods for online learning and stochastic
  optimization.
\newblock \emph{Journal of machine learning research}, 12\penalty0 (7), 2011.

\bibitem[Ester et~al.(1996)Ester, Kriegel, Sander, and Xu]{ester1996density}
Ester, M., Kriegel, H.-P., Sander, J., and Xu, X.
\newblock A density-based algorithm for discovering clusters in large spatial
  databases with noise.
\newblock In \emph{Proceedings of the Second International Conference on
  Knowledge Discovery and Data Mining}, pp.\  226--231, 1996.

\bibitem[Fang et~al.(2018)Fang, Li, Lin, and Zhang]{fang2018spider}
Fang, C., Li, C.~J., Lin, Z., and Zhang, T.
\newblock Spider: Near-optimal non-convex optimization via stochastic
  path-integrated differential estimator.
\newblock \emph{Advances in Neural Information Processing Systems}, 31, 2018.

\bibitem[Guo et~al.(2017{\natexlab{a}})Guo, Gao, Liu, and Yin]{guo2017improved}
Guo, X., Gao, L., Liu, X., and Yin, J.
\newblock Improved deep embedded clustering with local structure preservation.
\newblock In \emph{Ijcai}, pp.\  1753--1759, 2017{\natexlab{a}}.

\bibitem[Guo et~al.(2017{\natexlab{b}})Guo, Liu, Zhu, and Yin]{guo2017deep}
Guo, X., Liu, X., Zhu, E., and Yin, J.
\newblock Deep clustering with convolutional autoencoders.
\newblock In \emph{International conference on neural information processing},
  pp.\  373--382. Springer, 2017{\natexlab{b}}.

\bibitem[Hamerly \& Elkan(2003)Hamerly and Elkan]{hamerly2003learning}
Hamerly, G. and Elkan, C.
\newblock Learning the k in k-means.
\newblock \emph{Advances in neural information processing systems}, 16, 2003.

\bibitem[Hardt et~al.(2016)Hardt, Recht, and Singer]{hardt2016train}
Hardt, M., Recht, B., and Singer, Y.
\newblock Train faster, generalize better: Stability of stochastic gradient
  descent.
\newblock In \emph{International conference on machine learning}, pp.\
  1225--1234. PMLR, 2016.

\bibitem[Hoover et~al.(2022)Hoover, Chau, Strobelt, and
  Krotov]{hooveruniversal}
Hoover, B., Chau, D.~H., Strobelt, H., and Krotov, D.
\newblock A universal abstraction for hierarchical hopfield networks.
\newblock In \emph{The Symbiosis of Deep Learning and Differential Equations
  II}, 2022.

\bibitem[Hoover et~al.(2023)Hoover, Liang, Pham, Panda, Strobelt, Chau, Zaki,
  and Krotov]{hoover2023energy}
Hoover, B., Liang, Y., Pham, B., Panda, R., Strobelt, H., Chau, D.~H., Zaki,
  M.~J., and Krotov, D.
\newblock Energy transformer.
\newblock \emph{arXiv preprint arXiv:2302.07253}, 2023.

\bibitem[Hopfield(1982)]{hopfield1982neural}
Hopfield, J.~J.
\newblock Neural networks and physical systems with emergent collective
  computational abilities.
\newblock \emph{Proceedings of the national academy of sciences}, 79\penalty0
  (8):\penalty0 2554--2558, 1982.

\bibitem[Hopfield(1984)]{hopfield1984neurons}
Hopfield, J.~J.
\newblock Neurons with graded response have collective computational properties
  like those of two-state neurons.
\newblock \emph{Proceedings of the national academy of sciences}, 81\penalty0
  (10):\penalty0 3088--3092, 1984.

\bibitem[Huang et~al.(2017)Huang, Liu, Van Der~Maaten, and
  Weinberger]{huang2017densely}
Huang, G., Liu, Z., Van Der~Maaten, L., and Weinberger, K.~Q.
\newblock Densely connected convolutional networks.
\newblock In \emph{Proceedings of the IEEE conference on computer vision and
  pattern recognition}, pp.\  4700--4708, 2017.

\bibitem[Hubert \& Arabie(1985)Hubert and Arabie]{hubert1985comparing}
Hubert, L. and Arabie, P.
\newblock Comparing partitions.
\newblock \emph{Journal of classification}, 2\penalty0 (1):\penalty0 193--218,
  1985.

\bibitem[Hull(1994)]{hull1994database}
Hull, J.~J.
\newblock A database for handwritten text recognition research.
\newblock \emph{IEEE Transactions on pattern analysis and machine
  intelligence}, 16\penalty0 (5):\penalty0 550--554, 1994.

\bibitem[Ji et~al.(2017)Ji, Zhang, Li, Salzmann, and Reid]{ji2017deep}
Ji, P., Zhang, T., Li, H., Salzmann, M., and Reid, I.
\newblock Deep subspace clustering networks.
\newblock \emph{Advances in neural information processing systems}, 30, 2017.

\bibitem[Johnson(1967)]{johnson1967hierarchical}
Johnson, S.~C.
\newblock Hierarchical clustering schemes.
\newblock \emph{Psychometrika}, 32\penalty0 (3):\penalty0 241--254, 1967.

\bibitem[Kim \& Park(2008)Kim and Park]{kim2008sparse}
Kim, J. and Park, H.
\newblock Sparse nonnegative matrix factorization for clustering.
\newblock Technical report, Georgia Institute of Technology, 2008.
\newblock URL
  \url{https://faculty.cc.gatech.edu/~hpark/papers/GT-CSE-08-01.pdf}.

\bibitem[Kingma \& Ba(2014)Kingma and Ba]{kingma2014adam}
Kingma, D.~P. and Ba, J.
\newblock Adam: A method for stochastic optimization.
\newblock \emph{arXiv preprint arXiv:1412.6980}, 2014.

\bibitem[Krotov(2021)]{krotov2021hierarchical}
Krotov, D.
\newblock Hierarchical associative memory.
\newblock \emph{arXiv preprint arXiv:2107.06446}, 2021.

\bibitem[Krotov(2023)]{krotov2023new}
Krotov, D.
\newblock A new frontier for hopfield networks.
\newblock \emph{Nature Reviews Physics}, pp.\  1--2, 2023.

\bibitem[Krotov \& Hopfield(2016)Krotov and Hopfield]{krotov2016dense}
Krotov, D. and Hopfield, J.~J.
\newblock Dense associative memory for pattern recognition.
\newblock \emph{Advances in neural information processing systems}, 29, 2016.

\bibitem[Krotov \& Hopfield(2021)Krotov and Hopfield]{krotov2021large}
Krotov, D. and Hopfield, J.~J.
\newblock Large associative memory problem in neurobiology and machine
  learning.
\newblock In \emph{International Conference on Learning Representations}, 2021.

\bibitem[Li et~al.(2017)Li, Cheng, Wang, Morstatter, Trevino, Tang, and
  Liu]{li2017feature}
Li, J., Cheng, K., Wang, S., Morstatter, F., Trevino, R.~P., Tang, J., and Liu,
  H.
\newblock Feature selection: A data perspective.
\newblock \emph{ACM computing surveys (CSUR)}, 50\penalty0 (6):\penalty0 1--45,
  2017.

\bibitem[Liang et~al.(2022)Liang, Krotov, and Zaki]{liang2022modern}
Liang, Y., Krotov, D., and Zaki, M.~J.
\newblock Modern hopfield networks for graph embedding.
\newblock \emph{Frontiers in big Data}, 5, 2022.

\bibitem[Lloyd(1982)]{lloyd1982least}
Lloyd, S.
\newblock Least squares quantization in pcm.
\newblock \emph{IEEE transactions on information theory}, 28\penalty0
  (2):\penalty0 129--137, 1982.

\bibitem[Lucibello \& M{\'e}zard(2023)Lucibello and
  M{\'e}zard]{lucibello2023exponential}
Lucibello, C. and M{\'e}zard, M.
\newblock The exponential capacity of dense associative memories.
\newblock \emph{arXiv preprint arXiv:2304.14964}, 2023.

\bibitem[MacQueen(1967)]{macqueen1967classification}
MacQueen, J.
\newblock Classification and analysis of multivariate observations.
\newblock In \emph{5th Berkeley Symp. Math. Statist. Probability}, pp.\
  281--297, 1967.

\bibitem[March et~al.(2010)March, Ram, and Gray]{march2010fast}
March, W.~B., Ram, P., and Gray, A.~G.
\newblock Fast euclidean minimum spanning tree: algorithm, analysis, and
  applications.
\newblock In \emph{Proceedings of the 16th ACM SIGKDD international conference
  on Knowledge discovery and data mining}, pp.\  603--612, 2010.

\bibitem[McEliece et~al.(1987)McEliece, Posner, Rodemich, and
  Venkatesh]{mceliece1987capacity}
McEliece, R., Posner, E., Rodemich, E., and Venkatesh, S.
\newblock The capacity of the hopfield associative memory.
\newblock \emph{IEEE transactions on Information Theory}, 33\penalty0
  (4):\penalty0 461--482, 1987.

\bibitem[Millidge et~al.(2022)Millidge, Salvatori, Song, Lukasiewicz, and
  Bogacz]{millidge2022universal}
Millidge, B., Salvatori, T., Song, Y., Lukasiewicz, T., and Bogacz, R.
\newblock Universal hopfield networks: A general framework for single-shot
  associative memory models.
\newblock \emph{arXiv preprint arXiv:2202.04557}, 2022.

\bibitem[Nemirovski et~al.(2009)Nemirovski, Juditsky, Lan, and
  Shapiro]{nemirovski2009robust}
Nemirovski, A., Juditsky, A., Lan, G., and Shapiro, A.
\newblock Robust stochastic approximation approach to stochastic programming.
\newblock \emph{SIAM Journal on optimization}, 19\penalty0 (4):\penalty0
  1574--1609, 2009.

\bibitem[Nesterov(2003)]{nesterov2003introductory}
Nesterov, Y.
\newblock \emph{Introductory lectures on convex optimization: A basic course},
  volume~87.
\newblock Springer Science \& Business Media, 2003.

\bibitem[Panahi et~al.(2017)Panahi, Dubhashi, Johansson, and
  Bhattacharyya]{panahi2017clustering}
Panahi, A., Dubhashi, D., Johansson, F.~D., and Bhattacharyya, C.
\newblock Clustering by sum of norms: Stochastic incremental algorithm,
  convergence and cluster recovery.
\newblock In \emph{International conference on machine learning}, pp.\
  2769--2777. PMLR, 2017.
\newblock URL \url{http://proceedings.mlr.press/v70/panahi17a/panahi17a.pdf}.

\bibitem[Pedregosa et~al.(2011)Pedregosa, Varoquaux, Gramfort, Michel, Thirion,
  Grisel, Blondel, Prettenhofer, Weiss, Dubourg, et~al.]{pedregosa2011scikit}
Pedregosa, F., Varoquaux, G., Gramfort, A., Michel, V., Thirion, B., Grisel,
  O., Blondel, M., Prettenhofer, P., Weiss, R., Dubourg, V., et~al.
\newblock Scikit-learn: Machine learning in python.
\newblock \emph{the Journal of machine Learning research}, 12:\penalty0
  2825--2830, 2011.

\bibitem[Pelleg \& Moore(2000)Pelleg and Moore]{pelleg2000x}
Pelleg, D. and Moore, A.~W.
\newblock X-means: Extending k-means with efficient estimation of the number of
  clusters.
\newblock In \emph{ICML}, volume~1, pp.\  727--734, 2000.

\bibitem[Peng et~al.(2016)Peng, Xiao, Feng, Yau, and Yi]{peng2016deep}
Peng, X., Xiao, S., Feng, J., Yau, W.-Y., and Yi, Z.
\newblock Deep subspace clustering with sparsity prior.
\newblock In \emph{IJCAI}, pp.\  1925--1931, 2016.

\bibitem[Ram \& Gray(2011)Ram and Gray]{ram2011density}
Ram, P. and Gray, A.~G.
\newblock Density estimation trees.
\newblock In \emph{Proceedings of the 17th ACM SIGKDD international conference
  on Knowledge discovery and data mining}, pp.\  627--635, 2011.

\bibitem[Ram et~al.(2009)Ram, Lee, March, and Gray]{ram2009linear}
Ram, P., Lee, D., March, W., and Gray, A.
\newblock Linear-time algorithms for pairwise statistical problems.
\newblock \emph{Advances in Neural Information Processing Systems}, 22, 2009.

\bibitem[Ramsauer et~al.(2020)Ramsauer, Sch{\"a}fl, Lehner, Seidl, Widrich,
  Adler, Gruber, Holzleitner, Pavlovi{\'c}, Sandve,
  et~al.]{ramsauer2020hopfield}
Ramsauer, H., Sch{\"a}fl, B., Lehner, J., Seidl, P., Widrich, M., Adler, T.,
  Gruber, L., Holzleitner, M., Pavlovi{\'c}, M., Sandve, G.~K., et~al.
\newblock Hopfield networks is all you need.
\newblock \emph{arXiv preprint arXiv:2008.02217}, 2020.

\bibitem[Ren et~al.(2022)Ren, Pu, Yang, Xu, Li, Pu, Yu, and He]{ren2022deep}
Ren, Y., Pu, J., Yang, Z., Xu, J., Li, G., Pu, X., Yu, P.~S., and He, L.
\newblock Deep clustering: A comprehensive survey.
\newblock \emph{arXiv preprint arXiv:2210.04142}, 2022.

\bibitem[Reynolds(2009)]{reynolds2009gaussian}
Reynolds, D.~A.
\newblock Gaussian mixture models.
\newblock \emph{Encyclopedia of biometrics}, 741\penalty0 (659-663), 2009.

\bibitem[Rousseeuw(1987)]{rousseeuw1987silhouettes}
Rousseeuw, P.~J.
\newblock Silhouettes: a graphical aid to the interpretation and validation of
  cluster analysis.
\newblock \emph{Journal of computational and applied mathematics}, 20:\penalty0
  53--65, 1987.

\bibitem[Sander et~al.(1998)Sander, Ester, Kriegel, and Xu]{sander1998density}
Sander, J., Ester, M., Kriegel, H.-P., and Xu, X.
\newblock Density-based clustering in spatial databases: The algorithm gdbscan
  and its applications.
\newblock \emph{Data mining and knowledge discovery}, 2\penalty0 (2):\penalty0
  169--194, 1998.

\bibitem[Saxena et~al.(2017)Saxena, Prasad, Gupta, Bharill, Patel, Tiwari, Er,
  Ding, and Lin]{saxena2017review}
Saxena, A., Prasad, M., Gupta, A., Bharill, N., Patel, O.~P., Tiwari, A., Er,
  M.~J., Ding, W., and Lin, C.-T.
\newblock A review of clustering techniques and developments.
\newblock \emph{Neurocomputing}, 267:\penalty0 664--681, 2017.

\bibitem[Sibson(1973)]{sibson1973slink}
Sibson, R.
\newblock Slink: an optimally efficient algorithm for the single-link cluster
  method.
\newblock \emph{The computer journal}, 16\penalty0 (1):\penalty0 30--34, 1973.

\bibitem[Silverman(1986)]{silverman1986density}
Silverman, B.~W.
\newblock \emph{Density estimation for statistics and data analysis},
  volume~26.
\newblock CRC press, 1986.

\bibitem[Song et~al.(2013)Song, Liu, Huang, Wang, and Tan]{song2013auto}
Song, C., Liu, F., Huang, Y., Wang, L., and Tan, T.
\newblock Auto-encoder based data clustering.
\newblock In \emph{Iberoamerican congress on pattern recognition}, pp.\
  117--124. Springer, 2013.

\bibitem[Van~der Maaten \& Hinton(2008)Van~der Maaten and
  Hinton]{van2008visualizing}
Van~der Maaten, L. and Hinton, G.
\newblock Visualizing data using t-sne.
\newblock \emph{Journal of machine learning research}, 9\penalty0 (11), 2008.

\bibitem[Vassilvitskii \& Arthur(2006)Vassilvitskii and
  Arthur]{vassilvitskii2006k}
Vassilvitskii, S. and Arthur, D.
\newblock k-means++: The advantages of careful seeding.
\newblock In \emph{Proceedings of the eighteenth annual ACM-SIAM symposium on
  Discrete algorithms}, pp.\  1027--1035, 2006.

\bibitem[Vaswani et~al.(2017)Vaswani, Shazeer, Parmar, Uszkoreit, Jones, Gomez,
  Kaiser, and Polosukhin]{vaswani2017attention}
Vaswani, A., Shazeer, N., Parmar, N., Uszkoreit, J., Jones, L., Gomez, A.~N.,
  Kaiser, {\L}., and Polosukhin, I.
\newblock Attention is all you need.
\newblock \emph{Advances in neural information processing systems}, 30, 2017.

\bibitem[Vinh et~al.(2009)Vinh, Epps, and Bailey]{vinh2009information}
Vinh, N.~X., Epps, J., and Bailey, J.
\newblock Information theoretic measures for clusterings comparison: is a
  correction for chance necessary?
\newblock In \emph{Proceedings of the 26th annual international conference on
  machine learning}, pp.\  1073--1080, 2009.

\bibitem[Von~Luxburg(2007)]{von2007tutorial}
Von~Luxburg, U.
\newblock A tutorial on spectral clustering.
\newblock \emph{Statistics and computing}, 17\penalty0 (4):\penalty0 395--416,
  2007.

\bibitem[Xiao et~al.(2017)Xiao, Rasul, and Vollgraf]{xiao2017fashion}
Xiao, H., Rasul, K., and Vollgraf, R.
\newblock Fashion-mnist: a novel image dataset for benchmarking machine
  learning algorithms.
\newblock \emph{arXiv preprint arXiv:1708.07747}, 2017.

\bibitem[Xie et~al.(2016)Xie, Girshick, and Farhadi]{xie2016unsupervised}
Xie, J., Girshick, R., and Farhadi, A.
\newblock Unsupervised deep embedding for clustering analysis.
\newblock In \emph{International conference on machine learning}, pp.\
  478--487. PMLR, 2016.

\bibitem[Xu \& Wunsch(2005)Xu and Wunsch]{xu2005survey}
Xu, R. and Wunsch, D.
\newblock Survey of clustering algorithms.
\newblock \emph{IEEE Transactions on neural networks}, 16\penalty0
  (3):\penalty0 645--678, 2005.

\bibitem[Yang et~al.(2017)Yang, Fu, Sidiropoulos, and Hong]{yang2017towards}
Yang, B., Fu, X., Sidiropoulos, N.~D., and Hong, M.
\newblock Towards k-means-friendly spaces: Simultaneous deep learning and
  clustering.
\newblock In \emph{international conference on machine learning}, pp.\
  3861--3870. PMLR, 2017.

\bibitem[Zhang et~al.(2019)Zhang, Li, You, Qi, Zhang, Guo, and
  Lin]{zhang2019self}
Zhang, J., Li, C.-G., You, C., Qi, X., Zhang, H., Guo, J., and Lin, Z.
\newblock Self-supervised convolutional subspace clustering network.
\newblock In \emph{Proceedings of the IEEE/CVF conference on computer vision
  and pattern recognition}, pp.\  5473--5482, 2019.

\bibitem[Zhou et~al.(2022)Zhou, Xu, Zheng, Chen, Bu, Wu, Wang, Zhu, Ester,
  et~al.]{zhou2022comprehensive}
Zhou, S., Xu, H., Zheng, Z., Chen, J., Bu, J., Wu, J., Wang, X., Zhu, W.,
  Ester, M., et~al.
\newblock A comprehensive survey on deep clustering: Taxonomy, challenges, and
  future directions.
\newblock \emph{arXiv preprint arXiv:2206.07579}, 2022.

\end{thebibliography}
\bibliographystyle{icml2023}

\newpage
\appendix
\onecolumn

\section{Experimental Details} \label{asec:exp-details}

\subsection{Dataset details} \label{asec:exp-details:data}

To evaluate \clam, we conducted our experiments on ten standard benchmark data sets. The datasets are taken from various sources such as Yale from ASU feature selection repository\footnote{http://featureselection.asu.edu/}~\citep{li2017feature}, USPS from Kaggle\footnote{https://www.kaggle.com/datasets/bistaumanga/usps-dataset}~\citep{hull1994database}, Fashion-MNIST from Zalando\footnote{https://github.com/zalandoresearch/fashion-mnist}~\citep{xiao2017fashion}, GCM from~\citet{chakraborty2021automated} and the rest of the datasets from the UCI machine learning repository\footnote{https://archive.ics.uci.edu/ml/index.php} \citep{dua2017uci}. The statistics of datasets used in our experiment are given in Table \ref{tab:data}.

\begin{table}[htb]
\caption{Descriptions of various benchmark datasets, used in our experiments.}
\label{tab:data}
\begin{center}
\begin{tabular}{llccc}
\toprule
Dataset & Short name & \# Points & \# Features & \# Classes \\
\midrule
Zoo & Zoo &101 &16 &7 \\
Yale& Yale  &165 &1024 &15\\
GCM & GCM & 190 &16063 &14 \\
Ecoli & Ecoli & 336 &7 &8\\
Movement Libras & MovLib & 360 &90 &15\\
Mice Protien Expression & MicePE & 1080 &77 &8\\
USPS & USPS & 2007 &256 &10\\
CTG & CTG & 2126 &21 &10 \\
Segment & Segment & 2310 &19 &7 \\
Fashion MNIST & FMNIST & 60000 &784 &10\\
\bottomrule
\end{tabular}
\end{center}
\end{table}

\subsection{Metrics used} \label{asec:expt-details:metric}
To evaluate the performance of \clam, we use Silhouette Coefficient (SC)~\citep{rousseeuw1987silhouettes} as the unsupervised metric that is used to measure the quality of clustering. The score is between $-1$ and $1$; a value of $1$ indicates perfect clustering while a value of $-1$ indicates entirely incorrect clustering labels. A value of near 0 indicates that there exist overlapping clusters in the partition. To observe which existing clustering scheme is \clam most similar to, we use Normalized Mutual Information (NMI)~\citep{vinh2009information} \& Adjusted Rand Index (ARI)~\citep{hubert1985comparing} between the obtained partition by \clam and obtained partition by baselines. For NMI, a value of 1 indicates perfect clustering while a value of 0 indicates completely wrong class labels. ARI scores range between $-1$ and $1$ and the interpretation is same as SC. We also use the NMI \& ARI scores between the ground truth and the obtained partition from \clam and the baselines to measure how they are aligned to true clustering labels.

\subsection{Implementation Details} \label{asec:expt-details:implementation}
We use Tensorflow~\citep{abadi2016tensorflow} numerical machine learning library to implement and evaluate our model. We train \clam on a single node with 1 NVIDIA Geforce RTX 3090 (24GB RAM), and 8-core 3.5GHz Intel Core-i9 CPUs (32GB RAM). We train on the original data with masking where we utilize the distortion between the ground-truth masked value and completed pattern from \clam as the loss function to find the best model. In the forward pass, in each step, the feature vector is updated in such a way so that gradually it moves toward one of the stored memories. In the backward path, the memories are learned to minimize the loss. Hyperparameters are tuned for each dataset to find the best result. Full details of the hyperparameters used in our model are given in Table \ref{tab:hp_clam}. For the baseline schemes of $k$-means, spectral, and agglomerative, we use the implementation from {\tt scikit-learn}~\citep{pedregosa2011scikit} library and tune different hyperparameters to get the best results for each dataset. For DCEC~\citep{guo2017deep}, as it is based on convolutional autoencoders (CAE) and works with only image dataset, we evaluate on three image datasets to compare with \clam (we leverage their Tensorflow implementation\footnote{https://github.com/XifengGuo/DCEC}). For the soft-clustering part of DEC~\citep{xie2016unsupervised} (where they utilize KL divergence loss between predicted and target probability distribution), besides their $k$-means-initialized cluster centers, we also employ random-initialized cluster centers (DEC$^r$) to study how randomization works in DEC soft clustering network.
The description of used hyperparameters and their roles in the baseline schemes are given in Table~\ref{tab:hp_baselines}. 
\begin{table}[!ht]
\caption{Hyperparameters, their roles and range of values for \clam.}
\label{tab:hp_clam}
\begin{center}
\small
\begin{tabular}{l l}
\toprule
\multicolumn{1}{c}{\bf Hyperparameter} &\multicolumn{1}{c}{\bf Used Values} \\
\midrule
Inverse temperature, $\beta$ & [$10^{-5}$ - 5] \\
Number of layers, $T = \nicefrac{1}{\alpha} =\nicefrac{\tau}{dt}$  & [2-20] \\
\midrule
Batch size  & [8, 16, 32, 64, 128, 256] \\
Adam initial learning rate, $\epsilon$ & [$10^{-4}$, $10^{-3}$, $10^{-2}$, $10^{-1}$] \\
Reduce LR by factor & 0.8 \\
Reduce LR patience (epochs) & 5 \\
Minimum LR & $10^{-5}$ \\
Reduce LR loss threshold & $10^{-3}$ \\
Maximum Number of epochs & 200 \\
Number of restart & 10 \\
\midrule
Mask probability & [0.1, 0.12, 0.15, 0.2, 0.25, 0.3] \\
Mask value & [`mean', `min', `max'] \\
\bottomrule
\end{tabular}
\end{center}
\end{table}
\begin{table}[!ht]
\caption{Best hyperparameters for different datasets for \clam.
}
\label{tab:bhp}
\begin{center}
{\small
\begin{tabular}{lcccccccc}
\toprule
\multicolumn{1}{l}{\bf Dataset}  &\multicolumn{1}{c}{\bf Inverse temperature, $\beta$} &\multicolumn{1}{c}{\bf Layers, $T$} &\multicolumn{1}{c}{\bf Initial learning rate} &\multicolumn{1}{c}{\bf Batch size} &\multicolumn{1}{c}{\bf Mask probability} & \multicolumn{1}{c}{\bf Mask value}\\
\midrule
Zoo & 2.4 & 10 & 0.1 & 8 & 0.2 & 'mean' \\
Yale & 0.06 & 10 & 0.1 & 8 & 0.15 & 'mean' \\
GCM & 0.0004 & 12 & 0.1 & 8 & 0.15 & 'max'  \\
Ecoli & 0.095 & 12 & 0.1 & 16 & 0.15 & 'mean' \\
MLib
& 0.7 & 7 & 0.01 & 8 & 0.15 & 'mean' \\
MPE
& 0.1 & 5 & 0.2 & 8 & 0.15 & 'mean'\\
USPS & 0.1 & 5 & 0.001 & 16 & 0.15 & 'min' \\
CTG & 0.4 & 12 & 0.1 & 8 & 0.1 & 'mean' \\
Segment & 0.1 & 7 & 0.1 & 8 & 0.1 & 'mean' \\
FMNIST & 0.0005 & 5 & 0.01 & 256 & 0.1 & 'max' \\
\bottomrule
\end{tabular}
}
\end{center}
\end{table}

\section{Additional Experimental Results} \label{asec:add-exp}

\subsection{Hyperparameter Dependency for \clam} \label{asec:add-exp:clam-hpo}

\begin{figure}[!ht]
\centering
\begin{subfigure}{0.32\columnwidth}
\centering
\centerline{\includegraphics[width=\textwidth]{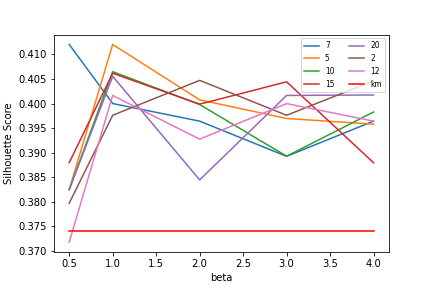}}
\caption{Zoo}
\label{fig:zoo_beta}
\end{subfigure}
~
\begin{subfigure}{0.32\columnwidth}
\centering
\centerline{\includegraphics[width=\textwidth]{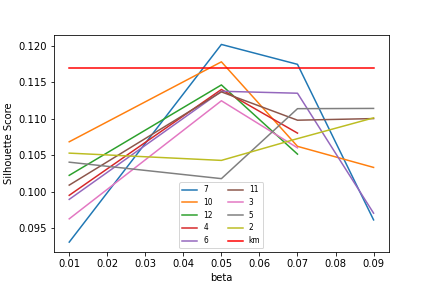}}
\caption{Yale}
\label{fig:movlib_beta}
\end{subfigure}
~
\begin{subfigure}{0.32\columnwidth}
\centering
\centerline{\includegraphics[width=\textwidth]{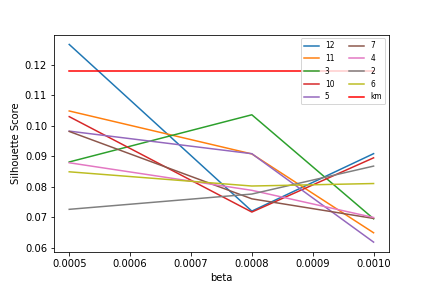}}
\caption{GCM}
\label{fig:mp_exp_beta}
\end{subfigure}
~
\begin{subfigure}{0.32\columnwidth}
\centering
\centerline{\includegraphics[width=\textwidth]{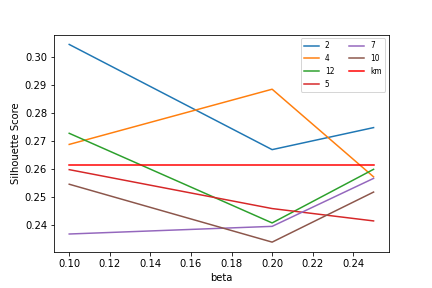}}
\caption{Ecoli}
\label{fig:usps_beta}
\end{subfigure}
~
\begin{subfigure}{0.32\columnwidth}
\centering
\centerline{\includegraphics[width=\textwidth]{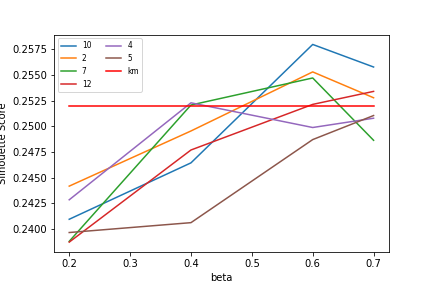}}
\caption{Movement Libras}
\label{fig:yale_beta}
\end{subfigure}
~
\begin{subfigure}{0.32\columnwidth}
\centering
\centerline{\includegraphics[width=\textwidth]{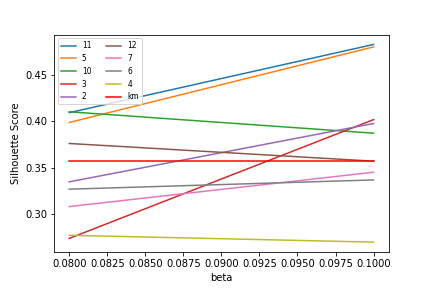}}
\caption{Segment}
\label{fig:ecoli_beta}
\end{subfigure}
\caption{{\bf Silhouette score vs inverse temperature ($\beta$) for six datasets and for different number of steps, $T$.} Label km refers to $k$-means.}
\label{fig:beta}
\end{figure}

\begin{figure}[!ht]
\centering
\begin{subfigure}{0.32\columnwidth}
\centering
\centerline{\includegraphics[width=\textwidth]{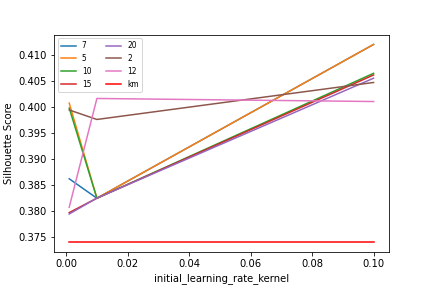}}
\caption{Zoo}
\label{fig:zoo_initial_learning_rate_kernel}
\end{subfigure}
~
\begin{subfigure}{0.32\columnwidth}
\centering
\centerline{\includegraphics[width=\textwidth]{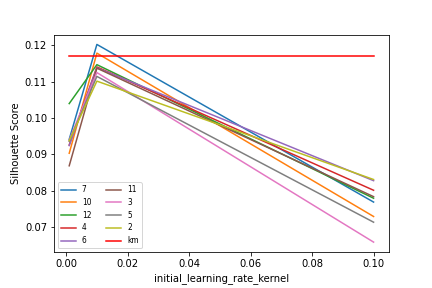}}
\caption{Yale}
\label{fig:movlib_initial_learning_rate_kernel}
\end{subfigure}
~
\begin{subfigure}{0.32\columnwidth}
\centering
\centerline{\includegraphics[width=\textwidth]{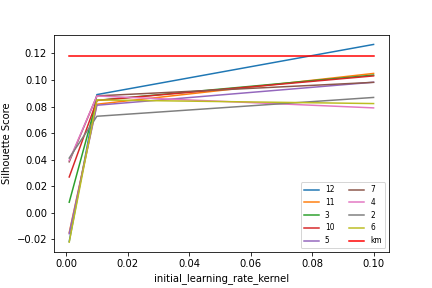}}
\caption{GCM}
\label{fig:mp_exp_initial_learning_rate_kernel}
\end{subfigure}
~
\begin{subfigure}{0.32\columnwidth}
\centering
\centerline{\includegraphics[width=\textwidth]{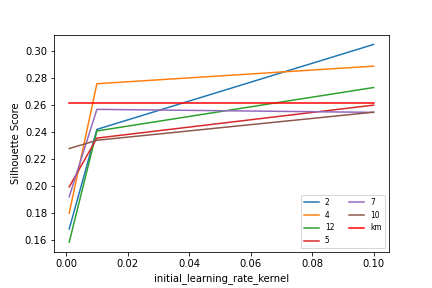}}
\caption{Ecoli}
\label{fig:usps_initial_learning_rate_kernel}
\end{subfigure}
~
\begin{subfigure}{0.32\columnwidth}
\centering
\centerline{\includegraphics[width=\textwidth]{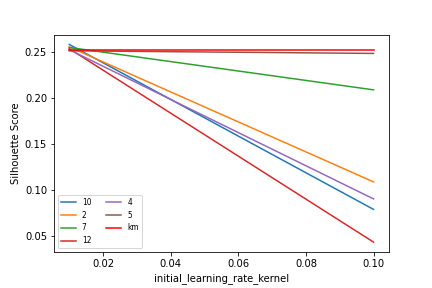}}
\caption{Movement Libras}
\label{fig:yale_initial_learning_rate_kernel}
\end{subfigure}
~
\begin{subfigure}{0.32\columnwidth}
\centering
\centerline{\includegraphics[width=\textwidth]{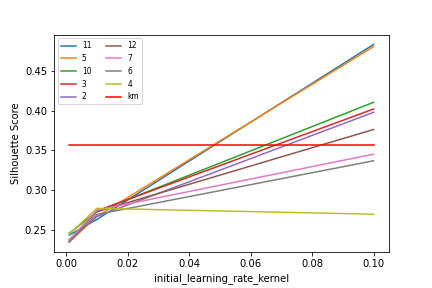}}
\caption{Segment}
\label{fig:ecoli_initial_learning_rate_kernel}
\end{subfigure}
\caption{{\bf Silhouette score vs initial learning rate for six datasets and for different number of steps, $T$.} Label km refers to $k$-means.}
\label{fig:irl}
\end{figure}

\begin{figure}[!htp]
\centering
\begin{subfigure}{0.32\columnwidth}
\centering
\centerline{\includegraphics[width=\textwidth]{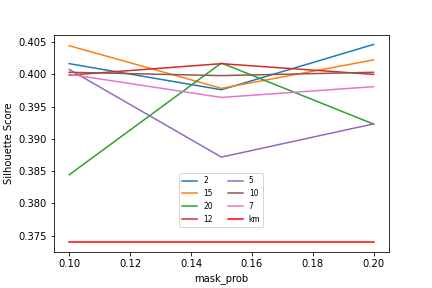}}
\caption{Zoo}
\label{fig:zoo_mask_prob}
\end{subfigure}
~
\begin{subfigure}{0.32\columnwidth}
\centering
\centerline{\includegraphics[width=\textwidth]{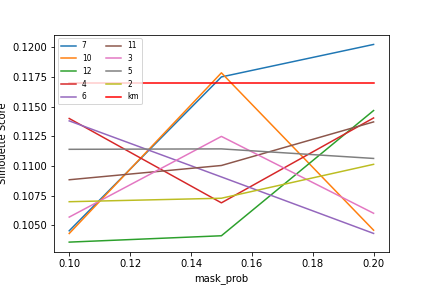}}
\caption{Yale}
\label{fig:movlib_mask_prob}
\end{subfigure}
~
\begin{subfigure}{0.32\columnwidth}
\centering
\centerline{\includegraphics[width=\textwidth]{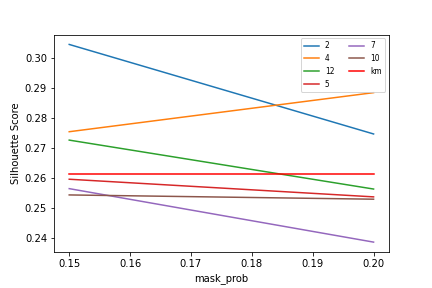}}
\caption{Ecoli}
\label{fig:usps_mask_prob}
\end{subfigure}
\caption{{\bf Silhouette score vs mask probability for three datasets and for different number of steps, $T$.} km means $k$-means.}
\label{fig:mask-prob}

\centering
\begin{subfigure}{0.32\columnwidth}
\centering
\centerline{\includegraphics[width=\textwidth]{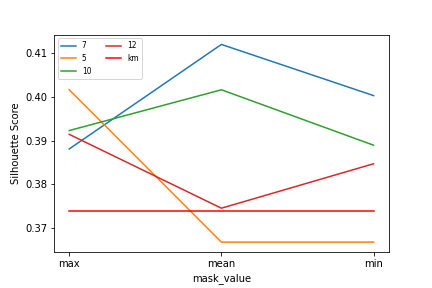}}
\caption{Zoo}
\label{fig:zoo_mask_value}
\end{subfigure}
~
\begin{subfigure}{0.32\columnwidth}
\centering
\centerline{\includegraphics[width=\textwidth]{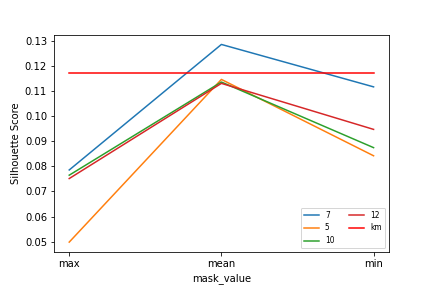}}
\caption{Yale}
\label{fig:movlib_mask_value}
\end{subfigure}
~
\begin{subfigure}{0.32\columnwidth}
\centering
\centerline{\includegraphics[width=\textwidth]{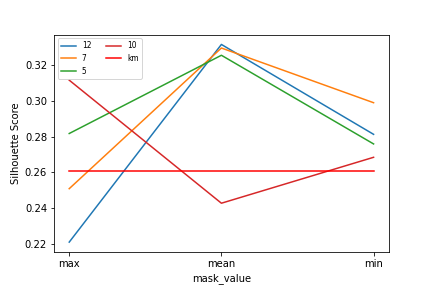}}
\caption{Ecoli}
\label{fig:usps_mask_value}
\end{subfigure}
\caption{{\bf Silhouette score vs mask values for three datasets and for different number of steps, $T$.} km is $k$-means.}
\label{fig:mask-value}
\end{figure}

To get the best result from \clam, we tune the involved hyperparameters (Table \ref{tab:hp_clam}) thoroughly. We use a range of [$10^{-5}- 2$] for the inverse temperature $\beta$, which is the most critical hyperparameter for \clam. Figure \ref{fig:beta} shows the effect of $\beta$ in measuring Silhouette Coefficient (SC)~\citep{rousseeuw1987silhouettes} for six datasets where each different plot indicates different number of steps ($T$) used in \clam. While SC is greatly dependent on the inverse temperature $\beta$ and steps $T$, we see that it is persistently competitive to the baseline $k$-means (red plot) for all the different configurations. We use the Adam optimizer and start with an initial learning rate, and we reduce the learning rate by a factor of 0.8 if the training loss does not ameliorate for a specific number of epochs until it reaches to the minimum learning rate threshold ($10^{-5}$). The effect of initial learning rate on \clam is shown in \figref{fig:irl}. We set the number of epochs to $200$ for each hyperparameter configuration, with number of restarts at $10$ (with different random seeds), and keep track of the training loss at the end of each epoch. We pick the set of hyperparameters and the related model for the inference step which produces the least training loss. For masking the original data, we tune different mask probabilities [$0.1 - 0.3]$ for different datasets to obtain the best model with three different mask values ('mean', 'min', 'max') for each feature. Figure \ref{fig:mask-prob} and \figref{fig:mask-value} depict the effect of mask probabilities and mask values on \clam, respectively. We can see that using 'mean' value of each feature as the mask value gives the nest results for almost all datasets. Table~\ref{tab:bhp} shows the best hyperparameters values for different datasets used in \clam.

\subsection{Hyperparameter Tuning for Baselines} \label{asec:expt-details:hpo}

We compare \clam with three baseline clustering schemes ($k$-means, spectral, and agglomerative) from {\tt scikit-learn}~\citep{pedregosa2011scikit}, DCEC~\citep{guo2017deep} \& soft DEC~\citep{xie2016unsupervised}. For $k$-means, spectral and agglomerative, we perform a comprehensive search for tuning different hyperparameters available in {\tt scikit-learn} and pick the best results. For DCEC and soft DEC, we use their suggested hyper-parameters with different update intervals (interval at which the predicted and target distributions get updated) to find the best results. Table \ref{tab:hp_baselines} provides a brief description of the hyperparameters and their roles in the baseline schemes.

\begin{table}[!h]
\caption{Hyperparameters (HPs), their roles and range of values for the baseline clustering schemes.}
\label{tab:hp_baselines}
\begin{center}
{\scriptsize
\begin{tabular}{l l| p{5.5cm}| p{3.5cm}}
\toprule
\multicolumn{1}{l}{\bf Baseline}  &\multicolumn{1}{c}{\bf HP} &\multicolumn{1}{c}{\bf Role} &\multicolumn{1}{c}{\bf Used Values} \\
\midrule
\multirow{4}{*}{$k$-means} & n\_clusters &  Number of clusters to be formed & True number of clusters\\
& init & Initialization method & [`k-means++', 'random'] \\
& n\_init & Number of time the k-means algorithm will be run  & 1000 \\
\midrule
\multirow{7}{*}{Spectral} & n\_clusters &  Number of clusters to be formed & True number of clusters\\
& affinity & The mechanism of constructing the affinity matrix & `nearest\_neighbors', `rbf' \\
& gamma & Kernel coefficient for `rbf', ignored for `nearest\_neighbors' & [0.001, 0.01, 0.05, 0.1, 0.5, 0.75, 1, 2, 5, 10] \\
& assign\_labels & Mechanism for assigning labels in the embedding space & [`$k$-means', `discretize'] \\
& n\_neighbors & Number of neighbors to consider when constructing the affinity matrix & [10, 15, 20, 50] \\
& n\_init & Number of time the k-means algorithm will be run  & 1000 \\
\midrule
\multirow{3}{*}{Agglomerative} & n\_clusters &  Number of clusters to be formed & True number of clusters\\
& affinity & Metric used to compute the linkage & [`euclidean', `l1', `l2', `manhattan', `cosine'] \\
& linkage & Linkage criterion to use & [`single',  `average', `complete', `ward'] \\
\midrule
\multirow{3}{*}{DCEC/DEC/DEC$^r$} & n\_clusters &  Number of clusters to be formed & True number of clusters\\
& batch\_size & Size of each batch & 256 \\
& maxiter & Maximum number of iteration & 2e4 \\
& gamma & Degree of freedom of student's t-distribution & 1 \\
& update\_interval & Interval at which the predicted and target distributions are updated & [1, 2, 3, 4, 5, 10, 20, 30, 40, 50, 75, 100, 125, 140, 150, 200] \\
& tol & Tolerance rate & 0.001 \\
\bottomrule
\end{tabular}
}
\end{center}
\end{table}

\subsection{Which clustering baseline is \clam most similar to?}\label{asec:emp:clam-similar}

We compare the ``overlap'' -- the normalized mutual information~\citep{vinh2009information} and adjusted rand index~\citep{hubert1985comparing} -- between the clusterings produced by \clam and the different baselines in Table~\ref{tab:amvsbasefull}.
As is evident from the results, the clusterings generated by \clam are significantly more aligned with those generated by $k$-means than the other baselines (with significant differences in 4/10 datasets). This is somewhat expected since both $k$-means and \clam are minimizing some form of the mean-squared-error (equivalently, the squared Euclidean distance), albeit in a different manner (alternating combinatorial optimization in $k$-means versus gradient descent based optimization in \clam).

\begin{table}[h]
\caption{NMI and ARI score comparison between the clusters found by \clam and the baseline clustering schemes.}
\label{tab:amvsbasefull}
\begin{center}
\small
\begin{tabular}{lcccc}
\toprule
\multicolumn{1}{l}{\bf Dataset}  &\multicolumn{1}{c}{\bf Metric} &\multicolumn{1}{c}{\bf $k$-means} &\multicolumn{1}{c}{\bf Spectral} &\multicolumn{1}{c}{\bf Agglomerative} \\
\midrule
\multirow{3}{*}{Zoo} & NMI & \textbf{0.8808} & 0.8315 & 0.8146 \\
& ARI & \textbf{0.7736} & 0.6691 & 0.6953  \\
\midrule 
\multirow{3}{*}{Yale} & NMI & \textbf{0.8846} & 0.8568 & 0.8840  \\
& ARI & \textbf{0.7870} & 0.7308 & 0.7621  \\
\midrule 
\multirow{3}{*}{GCM} & NMI & \textbf{0.4402} & 0.1400 & 0.1487 \\
& ARI & \textbf{0.1832} & -0.0223 & -0.0314  \\
\midrule 
\multirow{3}{*}{Ecoli} & NMI & 0.7448 & 0.7378 & \textbf{0.7718}  \\
& ARI & 0.5868 & 0.7381 & \textbf{0.8049}  \\
\midrule 
\multirow{3}{*}{Movement Libras} & NMI & \textbf{0.9158} & 0.7918 & 0.8426  \\
& ARI & \textbf{0.8110} & 0.6212 & 0.7009  \\
\midrule 
\multirow{3}{*}{Mice Protien Exp} & NMI & \textbf{0.4671} & 0.0043 & 0.4278  \\
& ARI & \textbf{0.3652} & 0.0003 & 0.2960  \\
\midrule 
\multirow{3}{*}{USPS} & NMI & \textbf{0.8150} & 0.1163 & 0.0227  \\
& ARI & \textbf{0.7869} & 0.0136 & 0.0023  \\
\midrule 
\multirow{3}{*}{CTG} & NMI & \textbf{0.5335} & 0.0449 & 0.0514  \\
& ARI & \textbf{0.3588} & 0.0106 & 0.0126  \\
\midrule 
\multirow{3}{*}{Segment} & NMI & \textbf{0.6359} & 0.0084 & 0.0.0070  \\
& ARI & \textbf{0.5484} & -0.0013 &  -0.0010  \\
\midrule 
\multirow{3}{*}{FMNIST} & NMI & \textbf{0.6391} & 0.5469 & 0.0023  \\
& ARI & \textbf{0.5182} & 0.4251 & 0.0021  \\
\bottomrule 
\end{tabular}
\end{center}
\end{table}

\subsection{Complete comparison to baselines in terms of ground truth} \label{asec:add-exp:baseline}

Table \ref{tab:fullgt} is a detailed version of Table \ref{tab:nmigt}, which shows the complete NMI and ARI score comparison between \clam and the baseline clustering schemes in terms of ground truth. From the table, we see that \clam not only performs well in terms of Silhouette Coefficient (SC), it performs equally well in terms of ground truth.

\begin{table}[!h]
\caption{NMI and ARI score comparison among \clam and the baseline clustering schemes in terms of ground truth.}
\label{tab:fullgt}
\begin{center}
\small
\begin{tabular}{lcccccccc}
\toprule
\multicolumn{1}{l}{\bf Dataset}  &\multicolumn{1}{c}{\bf Metric} &\multicolumn{1}{c}{\bf $k$-means} &\multicolumn{1}{c}{\bf Spectral} &\multicolumn{1}{c}{\bf Agglomerative} &\multicolumn{1}{c}{\bf DCEC} &\multicolumn{1}{c}{\bf DEC} &\multicolumn{1}{c}{\bf DEC$^r$} &\multicolumn{1}{c}{\bf \clam}\\
\midrule
\multirow{3}{*}{Zoo} & NMI & 0.8330 & 0.8891 & 0.8429 & N/A & 0.8330 & 0.7992 & \textbf{0.9429} \\
& ARI & 0.7373 & 0.8664 & 0.9109 & N/A & 0.7373 & 0.6755 & \textbf{0.9642} \\
\midrule
\multirow{3}{*}{Yale} & NMI & 0.6034 & 0.5744 & \textbf{0.6723} & 0.5428 & 0.5499 & 0.5068 & 0.6418 \\
& ARI & 0.3644 & 0.3226 & \textbf{0.4567} & 0.2817 & 0.2920 & 0.2398 & 0.4230 \\
\midrule
\multirow{3}{*}{GCM} & NMI & 0.4385 & 0.1715 & 0.1882 & N/A & 0.4228 & 0.3923 & \textbf{0.4476} \\
& ARI & 0.1308 & -0.0119 & -0.0071 & N/A & 0.1188 & 0.0880 & \textbf{0.2492} \\
\midrule
\multirow{3}{*}{Ecoli} & NMI & 0.6332 & 0.6606 & \textbf{0.7111} & N/A & 0.6332 & 0.5707 & 0.6633 \\
& ARI & 0.4997 & 0.6505 & \textbf{0.7261} & N/A & 0.4997 & 0.3893 & 0.7027 \\
\midrule
\multirow{3}{*}{Movement Libras} & NMI & 0.6044 & 0.6118 & 0.6086 & N/A & 0.5959 & 0.3147 & \textbf{0.6142} \\
& ARI & 0.3260 & 0.3236 & 0.3144 & N/A & 0.3147 & 0.2223 & \textbf{0.3351} \\
\midrule
\multirow{3}{*}{Mice Protien Exp} & NMI & 0.2373 & 0.0056 & 0.2596 & N/A & 0.2873 & 0.2951 & \textbf{0.3108} \\
& ARI & 0.1260 & 0.0019 & 0.1558 & N/A & 0.1756 & 0.1796 & \textbf{0.1652} \\
\midrule
\multirow{3}{*}{USPS} & NMI & 0.5368 & 0.0777 & 0.0180 & \textbf{0.6961} & 0.5376 & 0.4538 & 0.5566\\
& ARI & 0.4304 & -0.0033 & 0.0002 & \textbf{0.5907} & 0.4306 & 0.3226 & 0.4537 \\
\midrule
\multirow{3}{*}{CTG} & NMI & 0.3581 & 0.0391 & 0.0419 & N/A &  0.3507 & \textbf{0.3587} & 0.3154 \\
& ARI & 0.1780 & 0.0060 & 0.0078 & N/A & 0.1766 & \textbf{0.1818} & 0.1736 \\
\midrule
\multirow{3}{*}{Segment} & NMI & 0.5846 & 0.0102 & 0.0085 & N/A & 0.5853 & \textbf{0.6102} & 0.5489 \\
& ARI & 0.4607 & 0.0005 & 0.0003 & N/A & 0.4612 & \textbf{0.5038} & 0.4331 \\
\midrule
\multirow{3}{*}{FMNIST} & NMI & 0.5036 &  \textbf{0.6429} & 0.0051 & 0.5948 & 0.5008 & 0.3339 &0.5183 \\
& ARI & 0.3461 & \textbf{0.4307} & 0.0005 & 0.4113 & 0.3369 & 0.2279 & 0.3665 \\
\bottomrule
\end{tabular}
\end{center}
\end{table}

\subsection{Comprehensive experiment on basins of attraction of \clam vs Voronoi partition} \label{asec:add-exp:clam-voronoi}

\begin{figure*}[!ht]
\centering
\begin{subfigure}{0.18\columnwidth}
\centering
\centerline{\includegraphics[width=\columnwidth]{am-vs-voronoi/beta_0.001.png}}
\caption{$\beta$ = 0.001.}
\label{afig:beta_0.001}
\end{subfigure}
~
\begin{subfigure}{0.18\columnwidth}
\centering
\centerline{\includegraphics[width=\columnwidth]{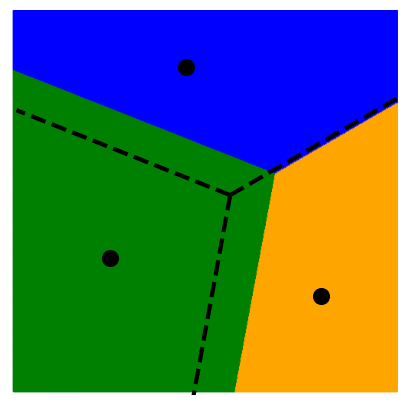}}
\caption{$\beta$ = 0.1.}
\label{afig:beta_0.1}
\end{subfigure}
~
\begin{subfigure}{0.18\columnwidth}
\centering
\centerline{\includegraphics[width=\columnwidth]{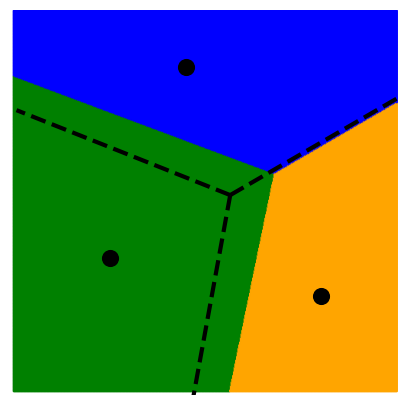}}
\caption{$\beta$ = 1.}
\label{afig:beta_1}
\end{subfigure}
~
\begin{subfigure}{0.18\columnwidth}
\centering
\centerline{\includegraphics[width=\columnwidth]{am-vs-voronoi/beta_10.png}}
\caption{$\beta$ = 10.}
\label{afig:beta_10}
\end{subfigure}
~
\begin{subfigure}{0.18\columnwidth}
\centering
\centerline{\includegraphics[width=\columnwidth]{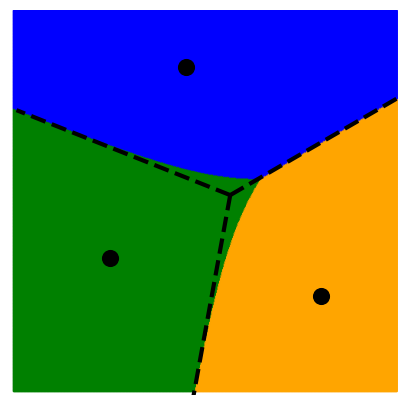}}
\caption{$\beta$ = 15.}
\label{afig:beta_15}
\end{subfigure}
~
\begin{subfigure}{0.18\columnwidth}
\centering
\centerline{\includegraphics[width=\columnwidth]{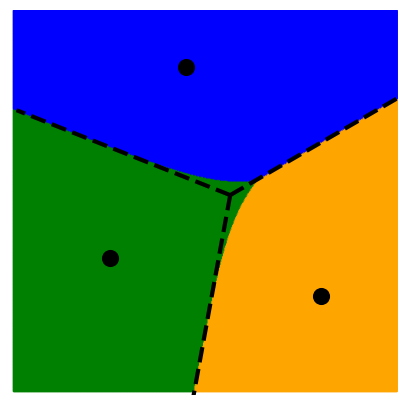}}
\caption{$\beta$ = 20.}
\label{afig:beta_20}
\end{subfigure}
~
\begin{subfigure}{0.18\columnwidth}
\centering
\centerline{\includegraphics[width=\columnwidth]{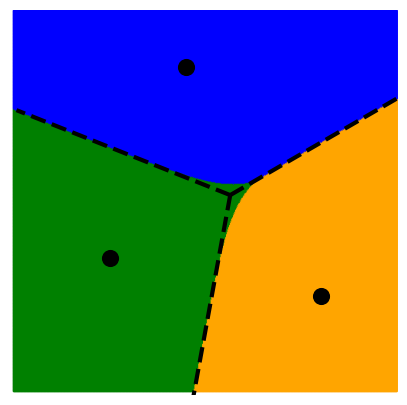}}
\caption{$\beta$ = 25.}
\label{afig:beta_25}
\end{subfigure}
~
\begin{subfigure}{0.18\columnwidth}
\centering
\centerline{\includegraphics[width=\columnwidth]{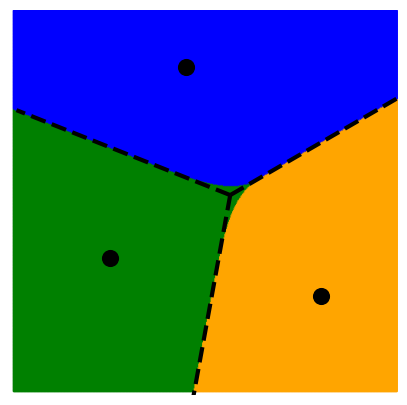}}
\caption{$\beta$ = 30.}
\label{afig:beta_30}
\end{subfigure}
~
\begin{subfigure}{0.18\columnwidth}
\centering
\centerline{\includegraphics[width=\columnwidth]{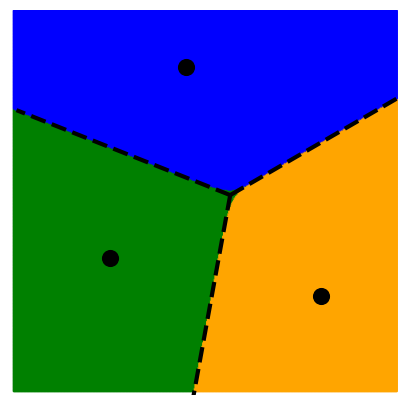}}
\caption{$\beta$ = 50.}
\label{afig:beta_50}
\end{subfigure}
~
\begin{subfigure}{0.18\columnwidth}
\centering
\centerline{\includegraphics[width=\columnwidth]{am-vs-voronoi/beta_100.png}}
\caption{$\beta$ = 100.}
\label{afig:beta_100}
\end{subfigure}
\caption{{\bf Basins of attraction vs Voronoi partition for three clusters.} Partitions induced by AM basins of attraction with given memories (black dots) for different $\beta$ are shown by the colored regions ($T$=10). Dashed lines show the Voronoi partition.
}
\label{fig:asec:add-exp:am-vs-voronoi}
\end{figure*}

\begin{figure*}[!ht]
\centering
\begin{subfigure}{0.18\columnwidth}
\centering
\centerline{\includegraphics[width=\columnwidth]{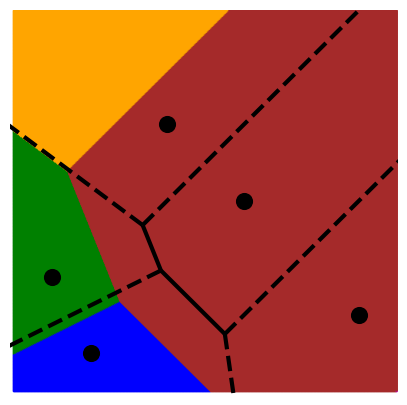}}
\caption{$\beta$ = 0.001.}
\label{afig:m5_beta_0.001}
\end{subfigure}
~
\begin{subfigure}{0.18\columnwidth}
\centering
\centerline{\includegraphics[width=\columnwidth]{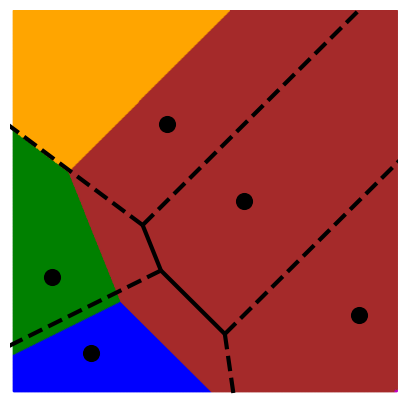}}
\caption{$\beta$ = 0.1.}
\label{afig:m5_beta_0.1}
\end{subfigure}
~
\begin{subfigure}{0.18\columnwidth}
\centering
\centerline{\includegraphics[width=\columnwidth]{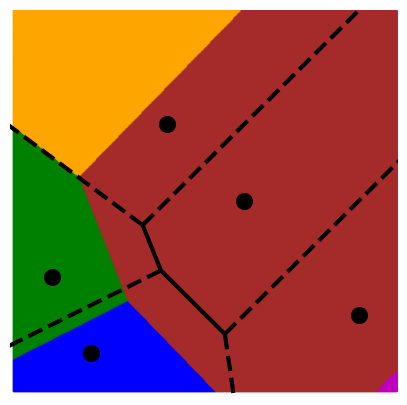}}
\caption{$\beta$ = 1.}
\label{afig:m5_beta_1}
\end{subfigure}
~
\begin{subfigure}{0.18\columnwidth}
\centering
\centerline{\includegraphics[width=\columnwidth]{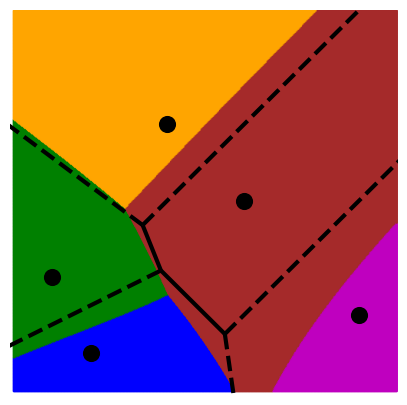}}
\caption{$\beta$ = 10.}
\label{afig:m5_beta_10}
\end{subfigure}
~
\begin{subfigure}{0.18\columnwidth}
\centering
\centerline{\includegraphics[width=\columnwidth]{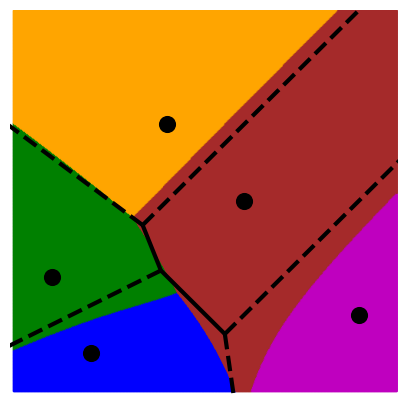}}
\caption{$\beta$ = 15.}
\label{afig:m5_beta_15}
\end{subfigure}
~
\begin{subfigure}{0.18\columnwidth}
\centering
\centerline{\includegraphics[width=\columnwidth]{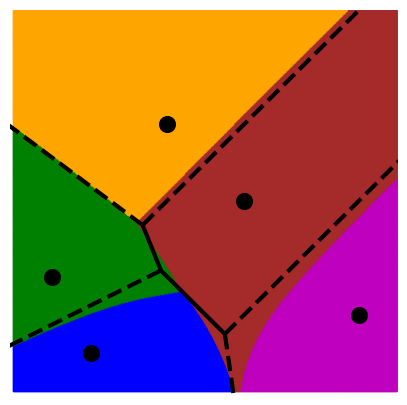}}
\caption{$\beta$ = 20.}
\label{afig:m5_beta_20}
\end{subfigure}
~
\begin{subfigure}{0.18\columnwidth}
\centering
\centerline{\includegraphics[width=\columnwidth]{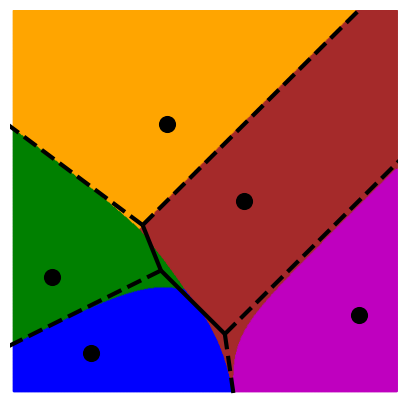}}
\caption{$\beta$ = 25.}
\label{afig:m5_beta_25}
\end{subfigure}
~
\begin{subfigure}{0.18\columnwidth}
\centering
\centerline{\includegraphics[width=\columnwidth]{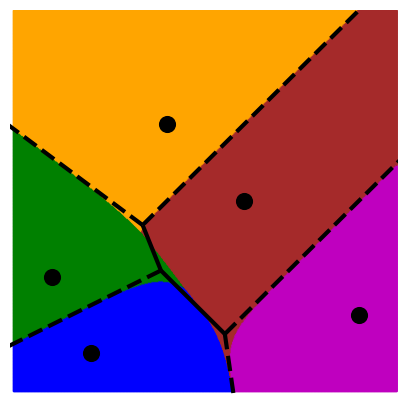}}
\caption{$\beta$ = 30.}
\label{afig:m5_beta_30}
\end{subfigure}
~
\begin{subfigure}{0.18\columnwidth}
\centering
\centerline{\includegraphics[width=\columnwidth]{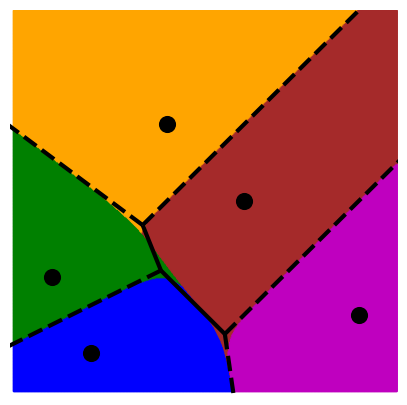}}
\caption{$\beta$ = 50.}
\label{afig:m5_beta_50}
\end{subfigure}
~
\begin{subfigure}{0.18\columnwidth}
\centering
\centerline{\includegraphics[width=\columnwidth]{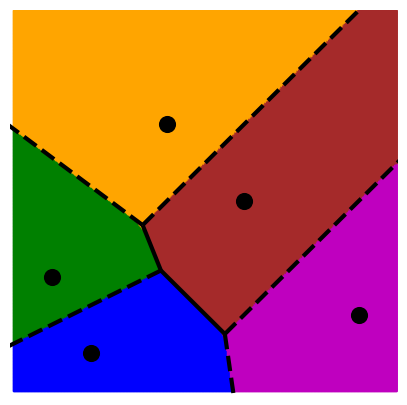}}
\caption{$\beta$ = 100.}
\label{afig:m5_beta_100}
\end{subfigure}
\caption{{\bf Basins of attraction vs Voronoi partition for five clusters.} Partitions induced by AM basins of attraction with given memories (black dots) for different $\beta$ are shown by the colored regions ($T$=10). Dashed lines show the Voronoi partition.
}
\label{fig:asec:add-exp:am-vs-voronoi-m5}
\end{figure*}

Figure~\ref{fig:asec:add-exp:am-vs-voronoi} (for three clusters) and Figure~\ref{fig:asec:add-exp:am-vs-voronoi-m5} (for five clusters) represent more comprehensive versions of experiments described in \figref{fig:am-vs-voronoi} to understand the evolution of AM basins of attraction to voronoi tesselation from low $\beta$ value $(0.001)$ to high $\beta$ value $(100)$ for step ($T$) $10$. From the figures, we see that starting with a completely non-voronoi partition ($\beta = 0.001)$, the basins of attraction of \clam follows the voronoi tesselation with some interesting non-linear characteristics ($\beta = 10-30)$, and then matches voronoi gradually ($\beta = 50-100)$. The experiment strongly indicates that \clam is equally able to find non-linear boundaries between the data points to find more compact partitions.

\subsection{Comprehensive experiment on elongated shaped clusters} \label{asec:add-exp:clam-elongated}
\begin{figure}[!ht]
\centering
\begin{subfigure}{0.225\columnwidth}
\centering
\centerline{\includegraphics[width=\textwidth]{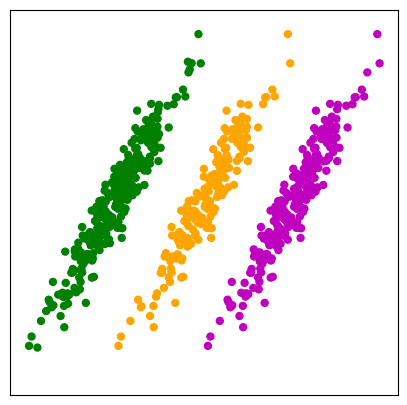}}
\caption{True}
\label{fig:3c_original}
\end{subfigure}
~
\begin{subfigure}{0.225\columnwidth}
\centering
\centerline{\includegraphics[width=\textwidth]{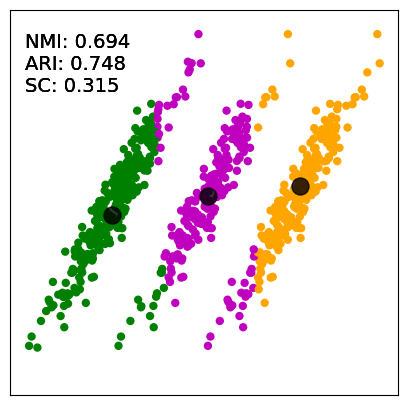}}
\caption{Lloyd's}
\label{fig:3c_kmeans}
\end{subfigure}
~
\begin{subfigure}{0.225\columnwidth}
\centering
\centerline{\includegraphics[width=\textwidth]{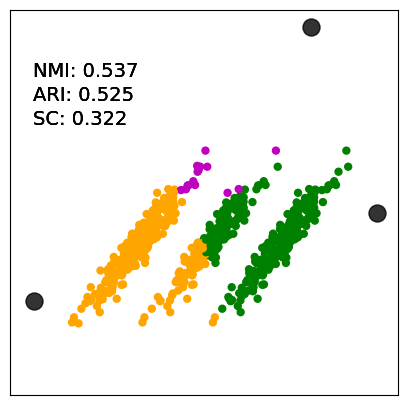}}
\caption{DEC}
\label{fig:3c_skmeans}
\end{subfigure}
~
\begin{subfigure}{0.225\columnwidth}
\centering
\centerline{\includegraphics[width=\textwidth]{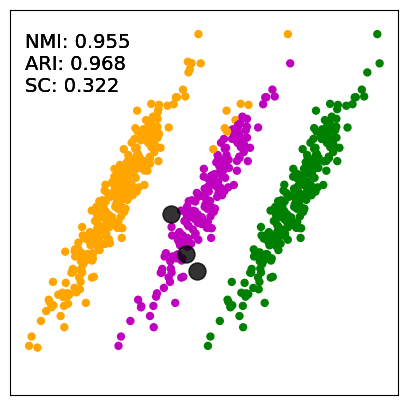}}
\caption{\clam}
\label{fig:3c_clam}
\end{subfigure}
\caption{{\bf Clustering with \clam.} Three clusters (\figref{fig:3c_original}), and solutions found by $k$-means~\citep{lloyd1982least} (\figref{fig:3c_kmeans}), DEC relaxation~\citep{xie2016unsupervised}(\figref{fig:3c_skmeans}), and our proposed end-to-end differentiable SGD-based \clam (\figref{fig:3c_clam}).
The black dots indicate the learned prototypes. {\em \clam discovers the ground-truth clusters while the baselines cannot}.
}
\label{fig:elongated}
\end{figure}

Figure \ref{fig:elongated} represents the further experiment, similar to that \figref{fig:teaser:toy-result}, for three elongated shaped clusters. Here, we can see \clam is able to find all three clusters in nearly perfect way where the baseline algorithms struggle to find the right partitions.

\subsection{Histogram of entropy of involved datasets} \label{asec:add-exp:entropy}

\begin{figure}[t]
\centering
\begin{subfigure}{0.32\columnwidth}
\centering
\centerline{\includegraphics[width=\textwidth]{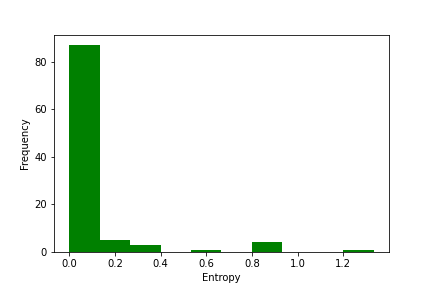}}
\caption{Zoo}
\label{fig:zoo}
\end{subfigure}
~
\begin{subfigure}{0.32\columnwidth}
\centering
\centerline{\includegraphics[width=\textwidth]{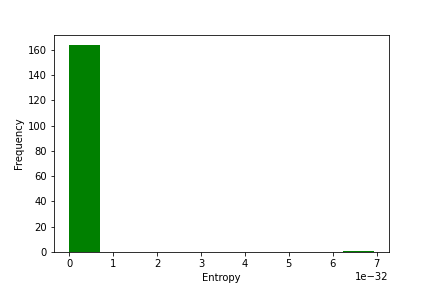}}
\caption{Yale}
\label{fig:yale}
\end{subfigure}
~
\begin{subfigure}{0.3\columnwidth}
\centering
\centerline{\includegraphics[width=\textwidth]{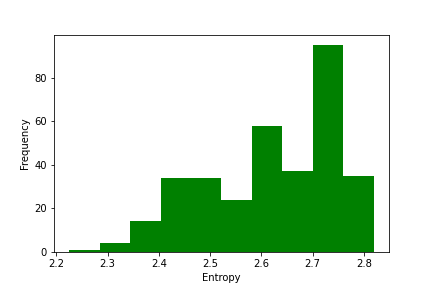}}
\caption{Ecoli}
\label{fig:ecoli}
\end{subfigure}
~
\begin{subfigure}{0.3\columnwidth}
\centering
\centerline{\includegraphics[width=\textwidth]{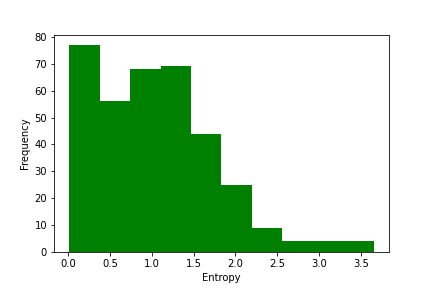}}
\caption{Movement Libras}
\label{fig:movlib}
\end{subfigure}
~
\begin{subfigure}{0.3\columnwidth}
\centering
\centerline{\includegraphics[width=\textwidth]{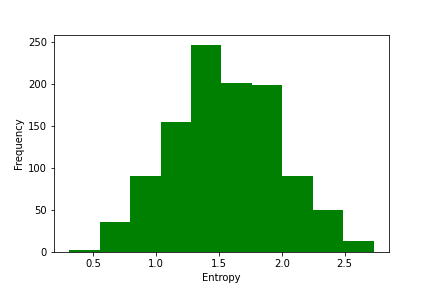}}
\caption{Mice Protien Expression}
\label{fig:mp_exp}
\end{subfigure}
~
\begin{subfigure}{0.3\columnwidth}
\centering
\centerline{\includegraphics[width=\textwidth]{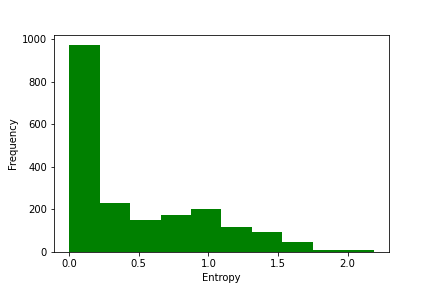}}
\caption{USPS}
\label{fig:usps}
\end{subfigure}
~
\begin{subfigure}{0.23\columnwidth}
\centering
\centerline{\includegraphics[width=\textwidth]{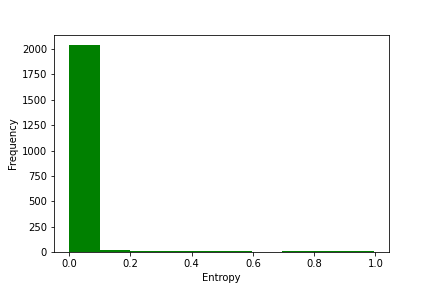}}
\caption{CTG}
\label{fig:ctg}
\end{subfigure}
~
\begin{subfigure}{0.23\columnwidth}
\centering
\centerline{\includegraphics[width=\textwidth]{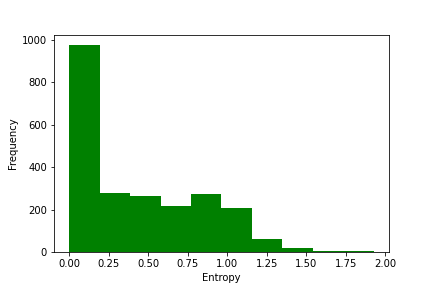}}
\caption{Segment}
\label{fig:segment}
\end{subfigure}
~
\begin{subfigure}{0.23\columnwidth}
\centering
\centerline{\includegraphics[width=\textwidth]{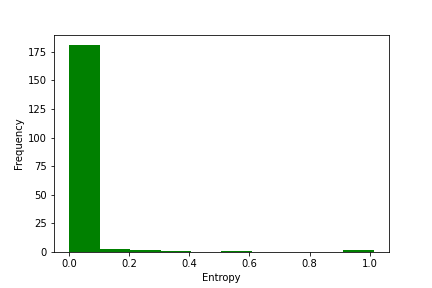}}
\caption{GCM}
\label{fig:gcm}
\end{subfigure}
~
\begin{subfigure}{0.23\columnwidth}
\centering
\centerline{\includegraphics[width=\textwidth]{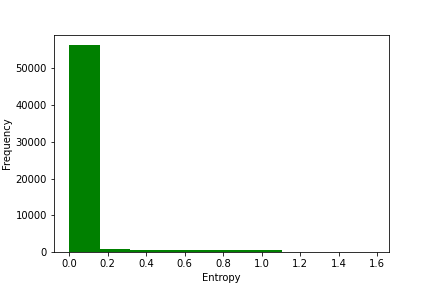}}
\caption{Fashion-MNIST}
\label{fig:fmnsit}
\end{subfigure}
\caption{\bf {Histogram of entropy for each of the ten datasets used to evaluate \clam.}}.
\label{fig:histogram}
\end{figure}

Figure \ref{fig:histogram} shows the histograms of entropy for each of the the ten datasets used to evaluate \clam. Equation \ref{eq:entropy} denotes how entropy is calculated on every data point $\vx$ in the dataset where we use the optimum $\beta$ (from Table \ref{tab:bhp}) and learned memories ($\brho_\mu$) after training for each dataset. 
\begin{equation}\label{eq:entropy}
H(\vx) = - \sum_{\mu \in [M]} \bf{p_{\mu}(\vx)} \log (\bf{p_{\mu}(\vx)}) \quad\quad \text{where} \quad \bf{p_{\mu}(\vx)} = \softmax(- \beta \| \brho_\mu - \vx \|^2 )
\end{equation}
These histograms represent the sharpness of the softmax function that explains how much the partition found by \clam is like a Voronoi partition where a value of zero means it always assigns the points to its closest memories and exactly matches with Voronoi.

\subsection{How does \clam handle noise in the input data?}
Observe that \clam considers the same objective function as the traditional $k$-means combinatorial optimization problem, and utilizes a new novel relaxation that allows us to solve the problem in an end-to-end differentiable manner in place of the well-established iterative discrete Lloyd’s algorithm. Therefore, robustness of \clam would be tied to the robustness of the original $k$-means objective to noise, and will of course depend on the nature of the noise.

One relevant form of noise in the context of clustering is the presence of outliers, and it is known that $k$-means is generally sensitive to outliers given that the $k$-means objective is effectively penalizing variance, which is sensitive to outliers. Spectral clustering can also be sensitive to outliers. Given that, we expect \clam to be potentially sensitive to outliers in the data. If we instead consider a more robust clustering objective in \clam (for example, by replacing the squared Euclidean distance in equation (6) with a Manhattan distance), it would be more robust to outliers.

\begin{figure}[t]
\centering
\begin{subfigure}{0.3\columnwidth}
\centering
\centerline{\includegraphics[height=1.5in]{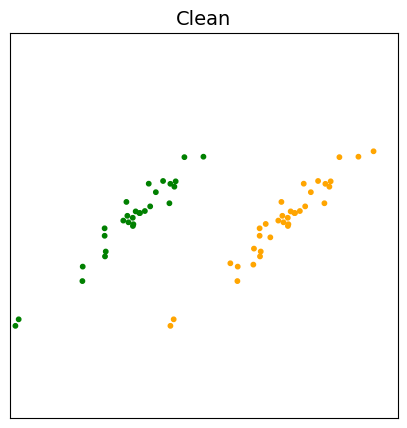}}
\caption{Clean}
\label{fig:clean}
\end{subfigure}
~
\begin{subfigure}{0.3\columnwidth}
\centering
\centerline{\includegraphics[height=1.5in]{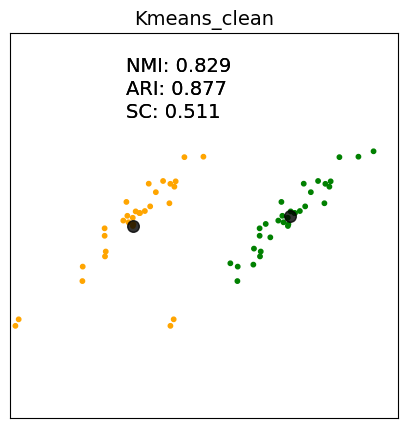}}
\caption{$k$-Means clean}
\label{fig:kmeans_clean}
\end{subfigure}
~
\begin{subfigure}{0.3\columnwidth}
\centering
\centerline{\includegraphics[height=1.5in]{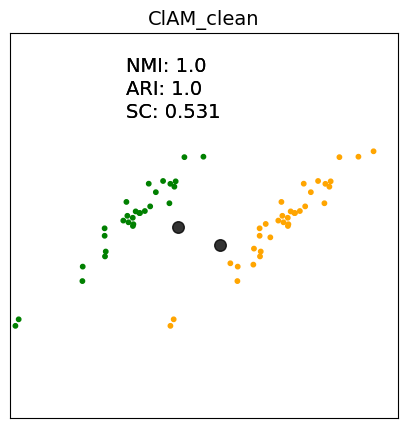}}
\caption{\clam clean}
\label{fig:clam_clean}
\end{subfigure}
~
\begin{subfigure}{0.3\columnwidth}
\centering
\centerline{\includegraphics[height=1.5in]{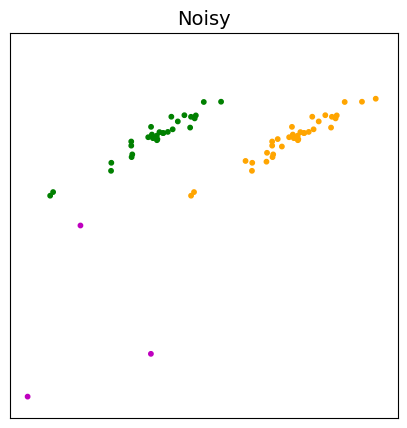}}
\caption{Noisy}
\label{fig:noisy}
\end{subfigure}
~
\begin{subfigure}{0.3\columnwidth}
\centering
\centerline{\includegraphics[height=1.5in]{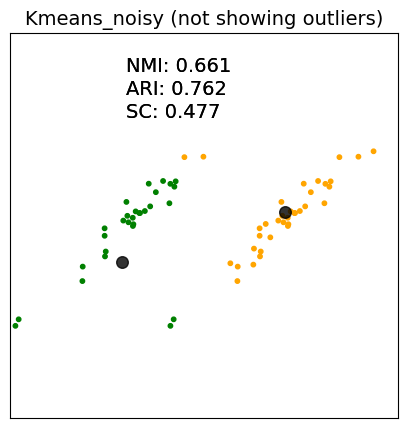}}
\caption{$k$-Means noisy}
\label{fig:kmeans_noisy}
\end{subfigure}
~
\begin{subfigure}{0.3\columnwidth}
\centering
\centerline{\includegraphics[height=1.5in]{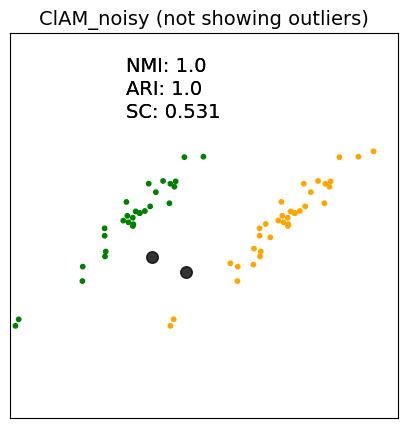}}
\caption{\clam noisy}
\label{fig:clam_noisy}
\end{subfigure}
\caption{{\bf Noise handling in input data:} top row -- clustering on original data, and bottom row -- clustering with outliers.}
\label{fig:noisy:toy-result}
\end{figure}

To evaluate the robustness to outliers, we consider a toy-example (Figure \ref{fig:noisy:toy-result}) with two clusters and we study the performance of $k$-means and \clam in the presence of outliers. Figure \ref{fig:clean} refers to the original dataset and Figure \ref{fig:noisy} refers to the dataset with outliers (which are shown in magenta color).
Next, we run $k$-means and \clam on both the original and noisy datasets (Figure \ref{fig:kmeans_clean}. Figure~\ref{fig:clam_clean} refers to the results on the original dataset, and Figure~\ref{fig:kmeans_noisy} and Figure \ref{fig:clam_noisy} refer to the results on the noisy dataset). We then compute the clustering quality via SC, NMI and ARI metrics. However, for the noisy dataset, we compute the metrics only on the inliers (i.e., excluding the outliers), to see how much the outliers affect the clustering quality of the actual inliers. Table~\ref{tab:noise} shows the performance of $k$-means and \clam as noise (outliers) is added to the data. The results indicate that, as expected, the addition of outliers reduces the clustering quality of $k$-means. However, \clam is able to cluster the data points correctly even with the outliers, with NMI and ARI both equal to 1.0. This indicates that \clam turns out to be robust to the outliers in this example. 

\begin{table}[t]
\caption{{\bf Noise handling in input data.}}
\label{tab:noise}
\begin{center}
{\footnotesize
\begin{tabular}{lcc}
\toprule
\multicolumn{1}{l}{\bf Method}  &\multicolumn{1}{c}{\bf SC (clean data)} &\multicolumn{1}{c}{\bf SC (noisy data)}\\
\midrule
$k$-means & 0.511 & 0.477 \\
\clam & 0.531 & 0.531 \\
\bottomrule
\end{tabular}
}
\end{center}
\end{table}

\vfill\pagebreak

\section{Technical details}

\subsection{Proof of Proposition~\ref{prop:train-time}}

Here we will provide a proof for Proposition~\ref{prop:train-time}. 

First consider the following optimization problem:
\begin{equation}\label{eq:opt-problem}
\min_\theta \sum_{\vx \in S} f(\vx; \theta)
\end{equation}

\begin{lemma}[\citet{nesterov2003introductory, fang2018spider}]\label{lem:sgd-sample-complexity}
For a smooth function $f$ (with respect to $\theta$), SGD converges to a $\epsilon$-stationary solution with $O|S|\epsilon^{-2})$ queries to the stochastic gradient oracle.
\end{lemma}

\begin{corollary}\label{cor:sgd-epochs}
SGD converges to a $\epsilon$-stationary solution with  $O(\epsilon^{-2})$ epochs over the dataset $S$.
\end{corollary}

We use these for proving Proposition~\ref{prop:train-time}:
\begin{proof}
For each point $\vx \in B$ in a batch, the computation cost of $T$ AM recursions, with each recursion taking $O(dk)$ time, takes a total of $O(dkT)$ time. Within one epoch, we perform the $T$ AM recursions for each $\vx \in S$, hence requiring $O(dkt |S|)$ time, where $|S|$ is the cardinality of $S$. Thus $N$ epochs in Train\clam takes $O(dkTN |S|)$ time.

If the objective in \eqref{eq:clam:masked} is smooth, then Corollary~\ref{cor:sgd-epochs} tells us that the optimization will converge to a $O(N^{-1/2})$-stationary point. So what remains to be shown is that the objective in \eqref{eq:clam:masked} is smooth, or more specifically, given $\mR = \{\brho_\mu, \mu \in [k] \}$ and $\mR' = \{ \brho_\mu', \mu \in [k]\}$, there exists a universal constant $C>0$
\begin{equation}\label{eq:clam-obj-smooth}
\E_{\vm \sim \mathcal M} \| \bar \vm \odot (\vx - \vx_\mR^T ) \|^2 - \E_{\vm \sim \mathcal M} \| \bar \vm \odot (\vx - \vx_{\mR'}^T ) \|^2
\leq C \sum_{\mu \in [k]} \| \brho_\mu - \brho_\mu' \|
\end{equation}
First, for binary masks $\vm \sim \mathcal M$
\begin{equation}
\E_{\vm \sim \mathcal M} \| \bar \vm \odot (\vx - \vx_\mR^T ) \|^2 - \E_{\vm \sim \mathcal M} \| \bar \vm \odot (\vx - \vx_{\mR'}^T ) \|^2 \leq \| \vx - \vx_\mR^T \|^2  - \| \vx - \vx_{\mR'}^T \|^2
\end{equation}
For $\vx \in S$ and bounded $\mR, \mR'$, triangle inequality gives us the following for some univeral constant $C_1$ such that
\begin{equation}\label{eq:obj-bound-treq}
\| \vx - \vx_\mR \|^2  - \| \vx - \vx_{\mR'} \|^2 \leq
C_1 \| \vx_\mR^T - \vx_{\mR'}^T \|
\end{equation}
thus, we need to show that $\| \vx_\mR^T - \vx_{\mR'}^T \| \leq C_2 \sum_{\mu \in [k]} \| \brho_\mu - \brho_\mu' \|$ for some universal constant $C_2$.

Now, we have from the AM recursion and Cauchy-Schwartz inequality that:
\begin{equation}\label{eq:particle-bound}
\| \vx_\mR^T - \vx_{\mR'}^T \| \leq \sum_{t = 1}^T \|\bdelta_\mR^t - \bdelta_{\mR'}^t\|
\end{equation}
where $\bdelta_\mR^t \triangleq \vx_\mR^t - \vx_\mR^{t-1}$, and $\bdelta_{\mR'}^t \triangleq \vx_{\mR'}^t - \vx_{\mR'}^{t-1}$.

Considering the first AM update $\bdelta_{\mR}^1, \bdelta_{\mR'}^1$, we have
\begin{align}
\|\bdelta_\mR^1 - \bdelta_{\mR'}^1\| 
& \leq 
\sum_{\mu \in [k]} \Big\| 
(\brho_\mu - \vx) \exp(-\beta \|\brho_\mu - \vx\|^2)
-
(\brho_\mu'- \vx) \exp(-\beta \|\brho_\mu' - \vx\|^2)
\Big\|
\\
& \leq 
\sum_{\mu \in [k]}\left(
\exp(-\beta \|\brho_\mu - \vx\|^2) \left\| 
\brho_\mu - \brho_\mu' 
\right\|
+
\|\brho_\mu' - \vx\|
\Big |
\exp(-\beta \|\brho_\mu - \vx\|^2)
-
\exp(-\beta \|\brho_\mu' - \vx\|^2)
\Big|
\right) \nonumber
\\
& \leq 
\sum_{\mu \in [k]}\left(
\left\| 
\brho_\mu - \brho_\mu' 
\right\|
+
\|\brho_\mu' - \vx\|
\Big |
\exp(-\beta \|\brho_\mu - \vx\|^2)
-
\exp(-\beta \|\brho_\mu' - \vx\|^2)
\Big|
\right) \nonumber
\\
& \leq C_3 \sum_{\mu \in [k]}  \left\| 
\brho_\mu - \brho_\mu' 
\right\|
\end{align}
for some universal $C_3$ for bounded $\vx, \brho_\mu, \brho_\mu'$ since the $\exp(\cdot)$ function is also smooth.

For the $(t+1)^{\text{th}}$ AM update $\bdelta_{\mR}^{t+1}, \bdelta_{\mR'}^{t+1}$, we have
\begin{align}
\|\bdelta_\mR^{t+1} - \bdelta_{\mR'}^{t+1}\| 
& \leq 
\sum_{\mu \in [k]} \Big\| 
(\brho_\mu - \vx_{\mR}^t) \exp(-\beta \|\brho_\mu - \vx_{\mR}^t\|^2)
-
(\brho_\mu'- \vx_{\mR'}^t) \exp(-\beta \|\brho_\mu' - \vx_{\mR'}^t\|^2)
\Big\|
\\
\begin{split}
& \leq 
\sum_{\mu \in [k]}
\left\|  \brho_\mu - \brho_\mu' \right\|
+
\left\|  \vx_{\mR}^t - \vx_{\mR'}^t \right\|
+
\|\brho_\mu' - \vx_{\mR'}^t\|
\left |
\exp(-\beta \|\brho_\mu - \vx_{\mR'}^t\|^2)
- 
\exp(-\beta \|\brho_\mu' - \vx_{\mR'}^t\|^2)
\right| \nonumber
\end{split}
\\
& \leq
C_3 \sum_{\mu \in [k]}  \left\| 
\brho_\mu - \brho_\mu' 
\right\|
+
k \left\|  \vx_{\mR}^t - \vx_{\mR'}^t \right\|
\\
& \leq
C_3 \sum_{\mu \in [k]}  \left\| 
\brho_\mu - \brho_\mu' 
\right\|
+
k \sum_{i=1}^t \left\|  \bdelta_{\mR}^i - \bdelta_{\mR'}^i \right\|
\end{align}

in the same way as with $\bdelta_{\mR}^1, \bdelta_{\mR'}^1$ except with the additional recursive sum $\sum_{i=1}^t \left\|  \bdelta_{\mR}^i - \bdelta_{\mR'}^i \right\|$. By induction, we can show that
\begin{align}
\|\bdelta_\mR^{t+1} - \bdelta_{\mR'}^{t+1}\| 
& \leq C_3 (k+1)^t
\sum_{\mu \in [k]}  \left\| 
\brho_\mu - \brho_\mu' 
\right\|
\end{align}
which we can apply to the RHS of \eqref{eq:particle-bound} to get the desired 
\begin{equation}\label{eq:particle-bound-complete}
\| \vx_\mR^T - \vx_{\mR'}^T \| \leq 
C_2
\sum_{\mu \in [k]}  \left\| 
\brho_\mu - \brho_\mu' 
\right\|
\end{equation}
for an universal constant $C_2$. This \eqref{eq:particle-bound-complete} combined with \eqref{eq:obj-bound-treq} gives us the desired condition in \eqref{eq:clam-obj-smooth} for smoothness of the \clam self-supervised loss function in \eqref{eq:clam:masked}. This completes the proof.
\end{proof}

\subsection{Proof of Proposition~\ref{prop:infer-time}}

\begin{proof}
The proof of Proposition~\ref{prop:infer-time} follows for combining the computation cost of $T$ AM recursions, with each recursion taking $O(dk)$ time, leading to a total of $O(dkT)$ time per point $\vx \in S$. Then the total runtime of Infer\clam is $O(dkT |S|)$, where $|S|$ is the cardinality of $S$.
\end{proof}
\end{document}